\pdfoutput=1
\PassOptionsToPackage{table}{xcolor}

\documentclass[11pt]{article}
\usepackage{pgfplots}
\pgfplotsset{compat=1.13}

\definecolor{col1}{HTML}{85929E}
\definecolor{col2}{HTML}{85C1E9}
\definecolor{col3}{HTML}{008080}

\usepackage{acl}
\newif\ifanon
\anonfalse

\usepackage{times}
\usepackage{latexsym}

\usepackage[T1]{fontenc}

\usepackage[utf8]{inputenc}

\usepackage{microtype}

\usepackage[scaled=.8]{beramono}

\usepackage{enumitem}

\usepackage{graphicx}
\usepackage[frozencache,cachedir=.]{minted}
\usepackage{booktabs}
\usepackage{algorithm}
\usepackage{algorithmic}
\usepackage{amsmath}
\usepackage{subcaption}
\usepackage{colortbl}
\usepackage{pifont}
\usepackage{colortbl}

\definecolor{class5}{RGB}{255, 204, 204} 
\definecolor{class4}{RGB}{255, 255, 204} 
\definecolor{class3}{RGB}{204, 255, 204} 
\definecolor{class2}{RGB}{204, 255, 255} 
\definecolor{class1}{RGB}{204, 204, 255} 
\definecolor{class0}{RGB}{255, 204, 255} 
\definecolor{NavyBlue}{RGB}{40, 160, 250} 

\newcommand{\sbs}[1]{%
    \pgfmathsetmacro{\normalized}{int((#1 - 0.4) / (1.0 - 0.4) * 100)}%
    \edef\tempa{\noexpand\cellcolor{NavyBlue!\normalized}}%
    \tempa#1%
}

\newcommand{\sbsalt}[1]{%
    \pgfmathsetmacro{\normalized}{int(min(100, max(0, #1 * 100)))}%
    \edef\tempa{\noexpand\cellcolor{NavyBlue!\normalized}}%
    \tempa#1%
}

\newcommand{\redx}{\cellcolor{red!40!}\ding{55}} 
\newcommand{\greentick}{\cellcolor{cyan!40!}\ding{51}} 

\usepackage[normalem]{ulem}

\usepackage[most]{tcolorbox}

\tcbset{
  parskip/.style={
    before={\par\pagebreak[0]\parindent=0pt},
    after={\parfillskip=0pt\par},
  },
}
\newtcolorbox{resultbox}[1][]{%
    colback=black!3,
    colframe=black!3,
    notitle,
    sharp corners,
    borderline west={2pt}{0pt}{gray!80!black},
    enhanced,
    breakable,
    boxsep=0pt,
    left=4pt,right=2pt,top=2pt,bottom=2pt,
    }

\newcommand{\dataset}{\texttt{mHumanEval}}

\newcommand{\he}{\texttt{HumanEval }}

\title{\dataset{} - A Multilingual Benchmark to Evaluate Large Language Models for Code Generation}

\author{Nishat Raihan, Antonios Anastasopoulos, Marcos Zampieri \\ George Mason University \\ Fairfax, VA, USA \\
\texttt{\{mraihan2, antonis, mzampier\}@gmu.edu}
}

\begin{document}

\maketitle
\begin{abstract}
Recent advancements in large language models (LLMs) have significantly enhanced code generation from natural language prompts. The \he Benchmark, developed by OpenAI, remains the most widely used code generation benchmark. However, this and other Code LLM benchmarks face critical limitations, particularly in task diversity, test coverage, and linguistic scope. Current evaluations primarily focus on English-to-Python conversion tasks with limited test cases, potentially overestimating model performance. While recent works have addressed test coverage and programming language (PL) diversity, code generation from low-resource language prompts remains largely unexplored. To address this gap, we introduce \dataset\footnote{\url{github.com/mraihan-gmu/mHumanEval-Benchmark}}, an extended benchmark supporting prompts in over 200 natural languages.
We employ established machine translation methods to compile the benchmark, coupled with a quality assurance process.
Furthermore, we provide expert human translations for 15 diverse natural languages (NLs). We conclude by analyzing the multilingual code generation capabilities of state-of-the-art (SOTA) Code LLMs, offering insights into the current landscape of cross-lingual code generation.
\end{abstract}

\section{Introduction} \label{intro}

LLMs have transformed software development with their ability to generate programming code from simple natural language instructions. LLMs are trained on extensive datasets that include diverse code samples, aiding programmers in code development and debugging. They also make programming more accessible to beginners. However, assessing the performance of these models across different coding tasks is still a major challenge. Comprehensive testing is essential to verify that these models are both effective and adaptable, rather than only performing well under specific conditions.

The most widely used benchmark for evaluating these models is OpenAI's \he \cite{chen2021codex}, which includes a collection of 164 tasks generated by human experts. Each task includes an English prompt, a canonical solution provided by the authors, and three test cases. Although this benchmark is commonly used, it has significant limitations, such as limited test coverage and minimal support for non-English and non-Python prompts. While recent variations \cite{peng2024humaneval, cassano2023multipl} of \he{}  address some of these issues, most do not include prompts in NLs other than English and, in particular, in low-resource NLs. Consequently, current benchmarks fail to provide key insights into the multilingual capabilities of LLMs in the context of code generation.

Figure \ref{fig:example} demonstrates one such example. While the widely used GPT3.5 \cite{brown2020language} model performs perfectly for the original prompt "\texttt{Write a Python code snippet that detects whether a year is a leap year or not.}", it fails when the same prompt is given in a low-resource language (\textit{Nyanja}, in this case).

\begin{figure}[!ht]
    \centering
    \small
    \begin{minipage}{\textwidth}
        \begin{minted}{python}
# Chaka chomwe tikufuna kuyang'ana
# Yang'anani ngati chaka ndi chaka cha ziwalo
if (year % 4 == 0 and year % 100 != 0):
    # Ngati chaka chimagawika ndi 4 ndipo 
    # sichimagawika ndi 100
    sindikiza(f"{year} ndi chaka cha ziwalo")
else:
    # Ngati sichigwirizana ndi zofunikira 
    # za chaka cha ziwalo
    sindikiza(f"{year} si chaka cha ziwalo")
        \end{minted}
    \end{minipage}
    \caption{\label{fig:example} Code snippet generated by GPT3.5 when prompted to write a Python code \texttt{to detect leap years} in \textit{Nyanja} language. Some Python keywords are transformed into \textit{Nyanja} words, resulting in compilation issues.}
\end{figure}

\noindent Most LLMs, primarily pre-trained on large English corpora like \href{https://commoncrawl.org/}{Common Crawl}, 
perform poorly on multilingual tasks, further propagating inequalities in language technology access~\cite{blasi-etal-2022-systematic}. However, proprietary models like GPT-4 \cite{openai2023gpt4} and Claude 3 \cite{anthropic2023claude3}, with undisclosed training data, show decent performance in multilingual scenarios. \citet{peng2024humaneval} for instance show that GPT-4 excels in code generation even with mid-resource language prompts. The open-source community is also advancing with multilingual models like Aya \cite{ustun2024aya} and \href{https://llama.meta.com/llama3/}{LLaMA 3}. However, insights into their code generation performance in a massively multilingual setting are lacking due to the absence of comprehensive benchmarks.

In this work, we introduce \dataset{}, a novel multilingual code generation benchmark including coding prompts in 204 NLs and expert human translations for 15 NLs. \dataset{} further includes canonical solutions in 25 PLs, including 4 new PLs that are not covered by any prior benchmarks. The primary contributions of this paper are as follows:

\begin{enumerate}
    \item The creation of \dataset{}, the first massively multilingual benchmark for code generation.
    \item A translation quality evaluation for each prompt. 
    \item A thorough evaluation of existing SOTA Code LLMs using \dataset{}.
\end{enumerate}

\noindent The paper addresses two research questions (RQs):

\begin{itemize}
    \item {\bf RQ1:} How do the code generation capabilities of LLMs vary when prompts are provided in English, or other high-, mid-, and low-resource NLs?
    \item {\bf RQ2:} How does the performance of multilingual LLMs compare to specialized, fine-tuned Code LLMs in code generation tasks on the \dataset{} dataset?
\end{itemize}


\vspace{2mm}

\noindent Finally, we also report \textit{secondary findings} related to the translation quality of machine translation (MT) methods on coding prompts. 



\section{Related Work}
\label{sec:rw}

The most widely used benchmark dataset for evaluating Code LLMs is the aformentioned \he \cite{chen2021codex}. Another key benchmark is DeepMind's \texttt{MBPP} \cite{austin2021program}, which includes 974 tasks with 3 test cases each. Despite their popularity, these benchmarks have significant limitations, such as inadequate test case coverage, limited number of PLs, and small task sets that do not represent real-world scenarios. Other benchmarks, like \texttt{CONCODE} \cite{iyer2018mapping} (Java), \texttt{AxiBench} \cite{hao2022aixbench} (Java), \texttt{CSEPrompts} \cite{raihan2024cseprompts} (Python) and \texttt{CodeApex} \cite{fu2023codeapex} (C++) focus on a single PL. 

To broaden PL coverage, \citet{cassano2023multipl} combine both \he and \texttt{MBPP} and add 17 more popular PLs besides Python, such as C++, Java, Ruby, and PHP. However, all prompts remain in English, with only 3 test cases per task. Similarly, the authors of \texttt{BabelCode} \cite{orlanski2023measuring} include 14 PLs and a more extensive test suite. To address test case coverage, \citet{liu2024your} introduce two datasets, \texttt{HumanEval+} and \texttt{MBPP+}, with significantly more test cases per task, ensuring both node and edge coverage. Notably, Code LLM performance decreases with the additional test cases, highlighting the initial benchmarks' limitations. Nevertheless, these benchmarks also use English prompts exclusively.

  

\begin{table*}[t]
\centering
\resizebox{0.8\textwidth}{!}{
\begin{tabular}{l|ccccc|c}
\toprule
 & \multicolumn{6}{c}{Benchmarks} \\
 & \texttt{HumanEval} & \texttt{MBPP} & \texttt{Babel} \texttt{Code} & \texttt{MultiPL-E} & \texttt{HumanEval-XL} & \texttt{mHumanEval} \\
\midrule
NL-Covg (MT) & 1 (\texttt{eng}) & 1 (\texttt{eng}) & 1 (\texttt{eng}) & 1 (\texttt{eng}) & 23 & \textbf{204} \\
NL-Covg (Human) & \redx & \redx & \redx & \redx & \redx & \textbf{15} \\
\midrule
PL-Covg & 1 (\texttt{py}) & 1 (\texttt{py}) & 14 & 18 & 12 & \textbf{25} \\
\bottomrule
\end{tabular}
}
\caption{Comparing popular benchmarks in terms of NL and PL coverage.}
\label{tab:comp}
\end{table*}

Few studies explore non-English coding prompts and evaluate Code LLMs on them. The recent benchmark, \texttt{HumanEval-XL} \cite{peng2024humaneval}, extends coverage for both NLs and PLs. This benchmark includes coding prompts in 23 NLs and solutions in 12 PLs. The original prompts from HumanEval \cite{chen2021codex} are translated into 23 different NLs using GPT-4 \cite{openai2023gpt4}, with the quality of these translations assessed using a thresholded \texttt{BERTScore} \cite{zhang2019BERTScore}. While \texttt{HumanEval-XL} explores multilingual prompts for code generation (Table \ref{tab:comp}), its 23 predominantly high-resource NLs limit insights into mid and low-resource NLs. The \textit{\texttt{BERTScore}} \cite{zhang2019BERTScore} evaluation may be inadequate, with \textit{\texttt{CometKiwi}} \cite{rei2023scaling} and \texttt{X-Comet} \cite{guerreiro2023xcomet} offering more robust alternatives. Experimenting with SOTA Code LLMs like WizardCoder \cite{luo2023wizardcoder} or multilingual models like Aya \cite{ustun2024aya} could yield valuable insights. Also, they do not include any human translations.

We argue that NL coverage is more critical than PL coverage when compiling a code generation benchmark. While prompts and tests can be reused across PLs, different NLs require curating contextually and linguistically appropriate prompts. Thus, NL diversity introduces more complexity in benchmark creation than PL diversity. To bridge the gap, we present \dataset{}, offering comprehensive experiments with multilingual coding prompts across 204 NLs and 25 PLs—the most extensive coverage to date (see Appendix \ref{app:list} for the full list) and the first one to include expert-human annotations (see Table \ref{tab:comp}). We describe \dataset{} in detail in this paper and we evaluate SOTA models on this dataset. 



\section{The \dataset{} Benchmark}

\begin{figure*}[t]
    \centering
    \includegraphics[width=.8\textwidth]{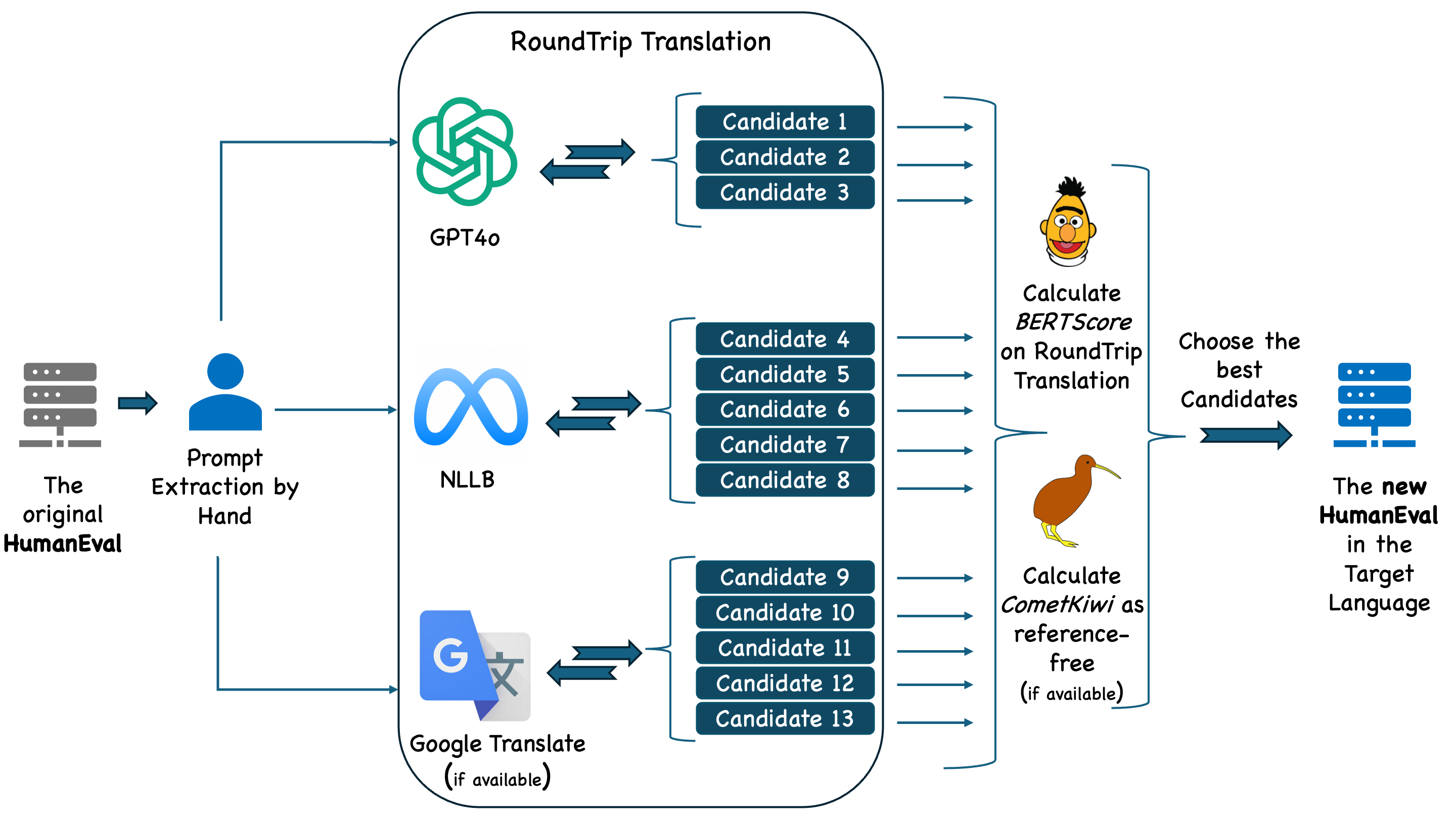}
    \caption{\label{fig:diagram} The workflow to generate prompts in a target language from the original \texttt{HumanEval}. Original prompts are first extracted manually. Then 3 Machine Translation models (GPT4o, NLLB, Google Translate) generate 13 candidates as well as roundtrip translations. Next, we evaluate each candidate's quality using \textit{\texttt{BERTScore}} using RoundTrip translations and \textit{\texttt{CometKiwi}} as a reference-free metric (if the language is supported). We then select the best candidate for each original prompt and compile the new benchmark for the target language.}
\end{figure*}

The \dataset{} benchmark is curated based on prompts from the original \he \cite{chen2021codex} dataset. It includes a total of 33,456 prompts, significantly expanding from the original 164. The curation process can be divided into several key steps, as illustrated in Figure \ref{fig:diagram} and elaborated upon in the following subsections.

\subsection{Prompt Extraction}
A typical prompt from the original dataset includes optional library imports, a function declaration, a \texttt{docstring}, and optional examples (as illustrated in Figure \ref{fig:example1}).

For translation, we only consider the \texttt{docstrings} (enclosed in triple quotes). These are manually extracted from all 164 prompts to ensure accuracy.

\begin{figure}[t]
    \centering
    \small
    \begin{minipage}{\textwidth}
        \begin{minted}{python}
        from typing import List 
        
        def all_prefixes: 
            """ Return list of all prefixes 
            from shortest to longest of the 
            input string. """
        
        >>> all_prefixes('abc')
        ['a', 'ab', 'abc']
        \end{minted}
    \end{minipage}
    \caption{\label{fig:example1}
    A sample prompt instance from the original \texttt{HumanEval} benchmark.}
\end{figure}

\subsection{Prompt Translation}

Upon extracting the prompts, we move on to translating them into different languages. We use three different machine translation strategies - leveraging OpenAI's \href{https://openai.com/index/hello-gpt-4o/}{GPT4-omni} 
through API, MetaAI's NLLB \cite{costa2022no}, which is the SOTA model for multiple NLs, and \href{https://cloud.google.com/translate/docs/reference/rest}{Google Translate} via API. 

Our target languages are all 204 languages from the Flores 200 dataset-\cite{costa2022no}. While we employ GPT4-omni and NLLB for all the target languages, it is important to note that we use only Google Translator for the 108 languages it supports (available through the API). 
For each extracted prompt, we employ the three translation systems for each target language, generating 5 candidate translation prompts (3 for GPT4o, due to budget considerations). We then evaluate the quality of the translation and keep the best one (see Figure \ref{fig:diagram}). The pseudocode is in Appendix~\ref{alg:eval}.

\begin{figure*}[t]
    \centering
    \includegraphics[width=0.725\textwidth]{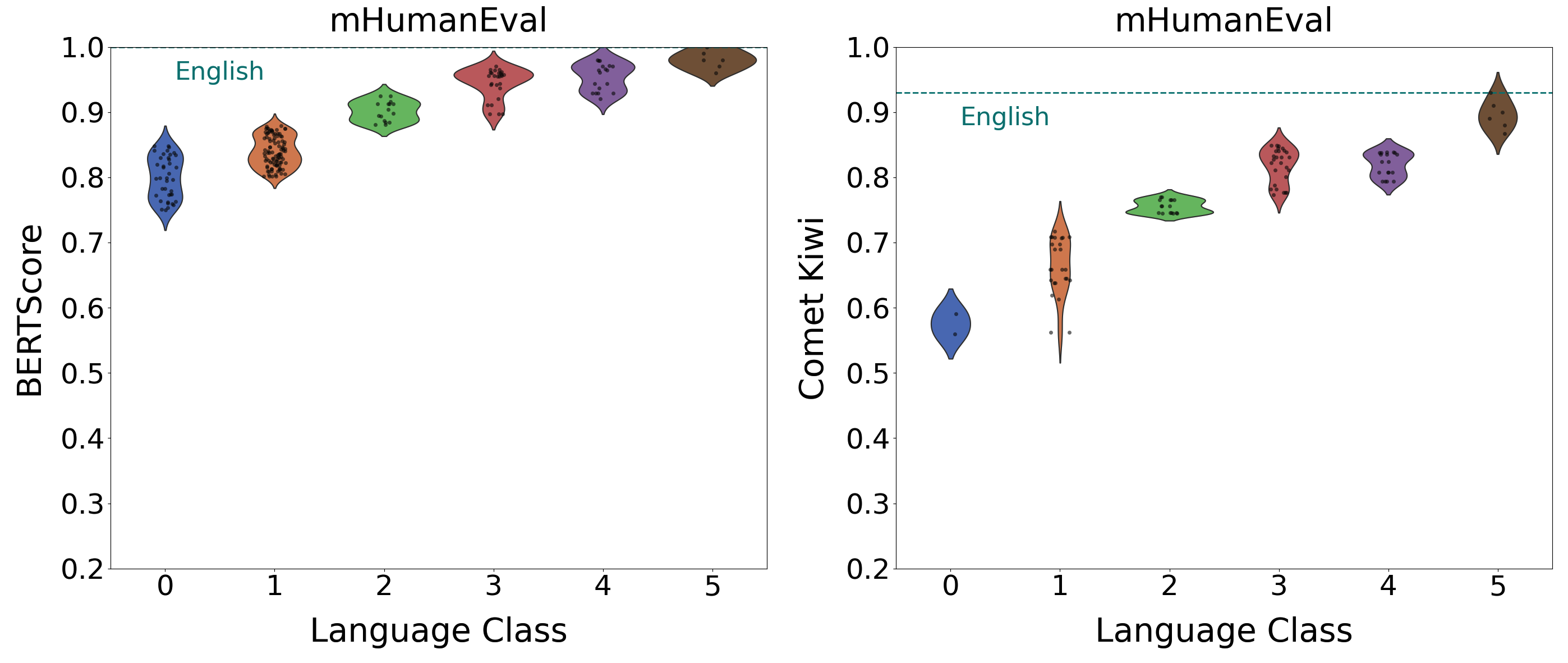}
    \caption{\label{fig:mt_mHumanEval} Evaluating the translated prompt qualities, chosen in \texttt{mHumanEval}. Our method results in better quality prompts compared to the model-specific translations (as depicted in Appendix \ref{app:comparison}).}
\end{figure*}

\subsection{Evaluating Prompt Quality}

We evaluate translation quality using \texttt{BERTScore} \cite{zhang2019BERTScore}, which focuses on similarity based on contextual embeddings, and \texttt{CometKiwi} \cite{rei2023scaling}, which is trained on human judgments of MT quality and incorporates linguistic features. 
While \texttt{BERTScore} uses BERT embeddings to measure candidate-reference translation similarity (Appendix \ref{app:bs}), \texttt{CometKiwi} evaluates translations reference-free, using human judgments and combining linguistic features with contextual embeddings (Appendix \ref{app:ck}). Using both ensures holistic evaluation, covering lexical similarity and human-assessed quality aspects.

As illustrated in Figure \ref{fig:diagram}, we generate 13 candidate translations for each prompt. We also perform round-trip translations back to the original language (eng\_Latn) to calculate the \texttt{BERTScore}. While \texttt{CometKiwi} is calculated as a reference-free metric. Both metrics generate scores in the \texttt{[0,1]} range. By computing the mean of the two metrics for each prompt, we select the candidate with the highest score. The mean scores for each language and system are provided in Appendices \ref{sec:mt_gpt4}, \ref{sec:mt_nllb} and \ref{sec:mt_gt}. It is worth noting that the \texttt{CometKiwi} metrics are not available for all languages, as it relies on XLM-R models \cite{conneau2019unsupervised,goyal2021larger} supporting 100 languages \cite{rei2023scaling}. For the remaining 104 Flores 200 languages, we use round-trip translations to calculate \texttt{BERTScore}, similar to \texttt{HumanEval-XL} \cite{peng2024humaneval}.

\subsection{Categorization based on Language Classes}
\label{sec:categories}

To better understand the performance of models on languages considered to be low- or high-resourced, we group the languages in \dataset{} following the methodology of \citet{joshi2020state}, who identify six classes of languages based on digital resource availability. These classes range from~0 to~5, with higher numbers indicating greater resource availability. \citeauthor{joshi2020state} classify a total of~2,485 languages, of which \dataset{} includes 204, including 15 with expert translations, as detailed in Table~\ref{tab:langs}.


\noindent We present the class-wise evaluation scores for the selected prompts in \texttt{mHumanEval} in Figure \ref{fig:mt_mHumanEval}. The language-specific scores are provided in Appendices \ref{sec:mt_gpt4}, \ref{sec:mt_nllb}, and \ref{sec:mt_gt}. Generally, the quality of the translation decreases as we address languages with fewer resources. However, by implementing Algorithm \ref{alg:translation_evaluation} and selecting from 13 candidate translations, the chosen candidates demonstrate improved quality compared to the model-specific results (see Appendices \ref{mt_mhumaneval} and \ref{app:comparison}). The final prompts in \dataset{} exhibit significantly better quality.

\begin{table}[!t]
    \small
  \centering
  \begin{tabular}{ccccc}
    \toprule
     \textbf{Class} & \textbf{Resource} & \textbf{Total} & \textbf{\dataset{}} & \textbf{\texttt{Expert}}  \\
    \midrule
    5 & High & 7 & 7 & 6\\
    4 & Mid to High & 18 & 18 & 4\\
    3 & Mid & 28 & 27 & 1\\
    2 & Low to Mid & 19 & 16 & 2\\
    1 & Low & 222 & 98 & 1\\
    0 & Rare & 2191 & 38 & 1\\
    \midrule
    ALL & -- & 2485 & 204 & 15\\
    \bottomrule
  \end{tabular}
  \caption{\label{tab:langs}
  Class distribution of natural languages based on resource availability. \textbf{Expert} denotes human translations done by expert programmers.}
\end{table}

\subsection{PL coverage}

As noted in Section \ref{sec:rw}, most benchmarks in this subdomain are limited to Python, including \he and \texttt{MBPP}. While recent benchmarks such as \texttt{MultiPL-E} and \texttt{HumanEval-XL} offer broader coverage, they still omit several widely used programming languages. With \dataset{}, we compile a comprehensive set of programming languages covered by existing multi-PL coding benchmarks and extend this set by incorporating four additional languages that have not been previously included: MATLAB, Visual Basic, Fortran, and COBOL (as shown in Table \ref{tab:benchmarks}).

We provide canonical solutions for the newly included four languages in the same format as \he{}. These solutions are handwritten by human experts and successfully pass all test cases. 

\subsection{\dataset{} Subsets}
We have a total of 33,456 prompts in \dataset{} spanning 204 NLs. Each prompt additionally supports 24 PLs, bringing the total number of prompts to 836,400. The entire dataset is publicly available on GitHub.

We also provide multiple subsets of the dataset for quick usability and interesting ablation studies (Table \ref{tab:mHumanEval}). Separate subsets are available for each NL and PL, in all possible combinations. Additionally, we create several variants for testing purposes- \texttt{mHumanEval-T500}: a subset consisting of the 500 highest-quality prompts based on \texttt{BERTScore} and \texttt{CometKiwi}, \texttt{mHumanEval-R500}: a randomly selected subset of 500 prompts, and \texttt{mHumanEval-B500}: a subset of the 500 lowest-quality prompts. Note that these prompts are drawn from the curated \dataset{}, which compiles the best prompts from 13 candidates each. Finally, we produce \texttt{mHumanEval-mini} which is a subset containing 204 prompts, with each prompt in a different language, where we select one prompt per language.

\begin{figure*}[t]
    \centering
    \includegraphics[width=0.85\textwidth]{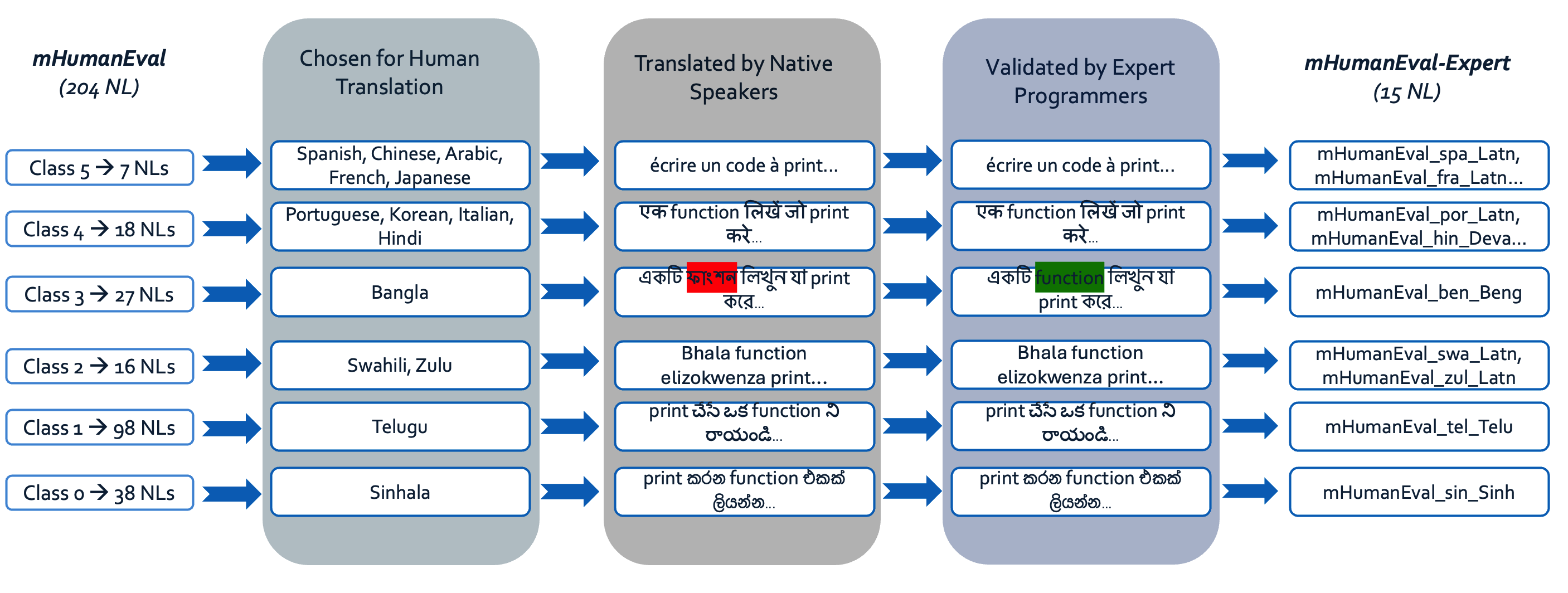}
    \caption{\label{fig:expert} Curating \dataset{}-Expert via native human translation followed by expert programmer evaluation.}
\end{figure*} 

\begin{table}[t]
  \centering
  \resizebox{0.4\textwidth}{!}{
  \begin{tabular}{lcc}
    \toprule
     & \textbf{Prompts} & \textbf{Note} \\
    \midrule
    \texttt{mHumanEval-\{NL\}}   & 164 each & Each NL\\
    \texttt{mHumanEval-mini}   & 204 & 204 NLs\\
    \texttt{mHumanEval-T500}   & 500 & Top 500\\
    \texttt{mHumanEval-R500}   & 500 & Random 500\\
    \texttt{mHumanEval-B500}   & 500 & Bottom 500\\
    \textbf{\texttt{mHumanEval-Expert}}   & 2460 & Human Generated\\
    \texttt{mHumanEval-\{PL\}}   & 4100 each & Each PL\\
    \textbf{\dataset{}}  & 33456 & Only Python \\
    \midrule
    \texttt{mHumanEval-Max}   & 836400 & All Prompts \\
    \bottomrule
  \end{tabular}}
  \caption{\label{tab:mHumanEval}
  Subsets and Variants of \dataset{}. These enable practitioners to carry out both comprehensive and preliminary evaluations on the benchmark.}
\end{table}

\subsection{\dataset{} - Expert}
\label{sec:expert}
The \dataset{}-Expert benchmark encompasses human translations across 15 languages, representing all six language classes (Table \ref{tab:langs}). Native speakers with computer science and engineering backgrounds perform these translations, ensuring precise interpretation of programming concepts and terminology. The curation process unfolds in three stages: (1) selection of 15 natural languages based on native speaker availability, ensuring representation from each language class; (2) translation by native speakers; and (3) quality assessment by expert programmers to verify the integrity of the coding prompts. Figure \ref{fig:expert} illustrates the whole curation process.

A comparative analysis between human translations and \dataset{}'s machine-translated prompts yields comparable evaluation metrics, with BERTScore variations of ±0.02 and CometKiwi variations of ±0.03 across the selected languages. Interestingly, annotators report no significant terminology concerns when reviewing machine translations. Further examination of the original \he prompts reveals that the docstrings—the primary translated content—predominantly comprise general task descriptions, minimizing the use of specialized coding terminology. This observation emphasizes the negligible discrepancies between human and machine translations in this context.

We conclude that human and machine translations of programming prompts across 15 languages show similar quality, with minimal differences in evaluation metrics. This similarity is attributed to the general nature of the content, which contains limited specialized coding terminology.

\section{Experiments}
\label{sec:eval}

\paragraph{Model Selection} We experiment with \dataset{} using six models (Table \ref{tab:models}), including both proprietary and open-source SOTA models for code generation. We use a mix of general-purpose and finetuned models to gather broader insights.

\begin{table}[!h]
\small
  \centering
  \scalebox{0.8}{
  \begin{tabular}{lccc}
    \toprule
    \textbf{Model} & \textbf{Size} & \textbf{Type} & \textbf{Ref.}\\
    \midrule
    GPT4o & -- & Base & \cite{openai2023gpt4} \\
    Claude-3.5-Opus & -- & Base & \cite{anthropic2023claude3} \\
    GPT3.5 & 175B & Base & \cite{brown2020language} \\
    DeepSeek-Coder-V2 & 236B & Finetuned & \cite{dai2024deepseekmoe} \\
    WizardCoder & 33B & Finetuned & \cite{luo2023wizardcoder} \\
    Aya & 33B & Finetuned & \cite{ustun2024aya} \\
    \bottomrule
  \end{tabular}}
  \caption{\label{tab:models}
  LLMs evaluated on \dataset{}.}
\end{table}


\paragraph{Prompting} We use the proprietary models through their APIs. Our experiments include all 33,456 prompts from \dataset{}, with 164 prompts for each language. We follow the standard prompt templates for each LLM. These templates are shown in Appendix~\ref{app:prompts}.

\paragraph{Code Execution} Following code generation, we move to execution. The six models produce well-structured code blocks, requiring minimal cleaning. We use simple RegEx commands to extract these blocks, and evaluate them locally in batches using Python's \texttt{subprocess}\footnote{\url{docs.python.org/3/library/subprocess.html}} library, focusing exclusively on the \textbf{\texttt{Pass@1}} metric.

\begin{figure*}[t]
    \centering
    \includegraphics[width=0.82\textwidth]{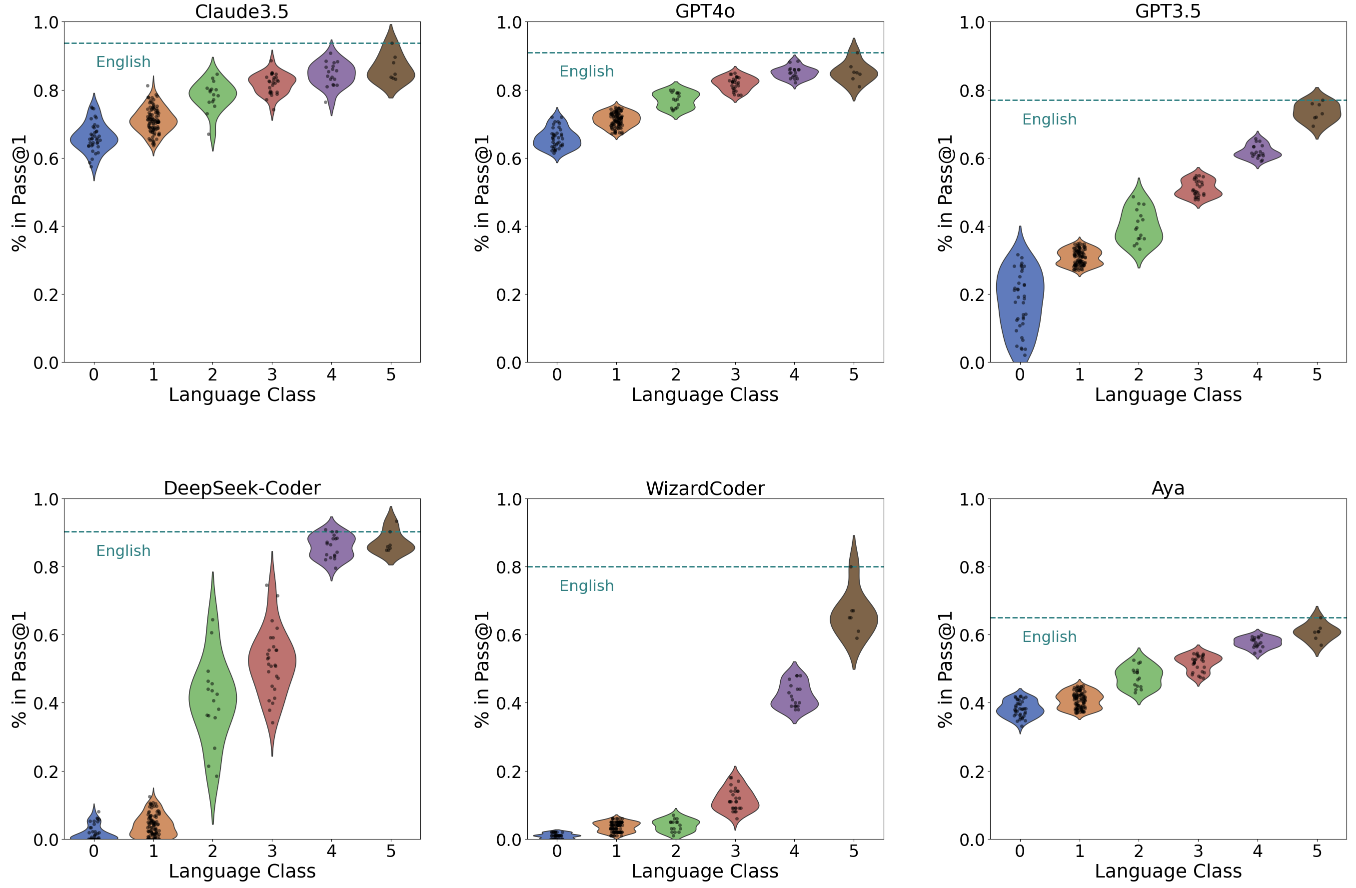}
    \caption{\label{fig:mHumanEval_plots_1} Comparing model performances (\% in \textbf{\texttt{Pass@1}}) for the six models on \dataset{}-\textbf{Python}.}
\end{figure*}

\paragraph{Results} For each language, we present the \textbf{\texttt{Pass@1}} scores as percentages, categorizing them by the six language classes as discussed in Section \ref{sec:categories}. As illustrated in Figure \ref{fig:mHumanEval_plots_1}, Claude3.5 and GPT4o exhibit the most consistent performance, maintaining strong results even with coding prompts in low-resource languages. In contrast, GPT3.5 and DeepSeek experience a significant decline in performance for low-resource classes. Although Aya shows the weakest results for higher resource classes, it maintains relative consistency, even in extremely low-resource languages. On the other hand, WizardCoder achieves excellent results in English and reasonable performance for Class 5, but its performance deteriorates significantly in other languages. The model and language-specific detailed results are presented in Appendix \ref{results}.

\paragraph{Other PLs} We extend our evaluation to four additional subsets of \dataset{}: \dataset{}-C++, \dataset{}-JAVA, \dataset{}-JavaScript, and \dataset{}-Ruby. The average \textbf{\texttt{Pass@1}} scores across all 204 NLs for the 5 PLs are shown in Table \ref{performance_summary}. 

\begin{table}[!h]
  \centering
  \resizebox{0.94\columnwidth}{!}{
  \begin{tabular}{lccccc}
    \toprule
     & \textbf{Python} & \textbf{Java} & \textbf{C++} & \textbf{JavaScript} & \textbf{Ruby} \\
    \midrule
    GPT4o           & \sbsalt{0.738} & \sbsalt{0.650} & \sbsalt{0.652} & \sbsalt{0.477} & \sbsalt{0.480} \\
    GPT3.5          & \sbsalt{0.360} & \sbsalt{0.270} & \sbsalt{0.270} & \sbsalt{0.099} & \sbsalt{0.103} \\
    Claude3.5       & \sbsalt{0.739} & \sbsalt{0.651} & \sbsalt{0.649} & \sbsalt{0.483} & \sbsalt{0.477} \\
    DeepSeek-Coder  & \sbsalt{0.229} & \sbsalt{0.139} & \sbsalt{0.136} & \sbsalt{0.000} & \sbsalt{0.000} \\
    WizardCoder     & \sbsalt{0.098} & \sbsalt{0.009} & \sbsalt{0.007} & \sbsalt{0.000} & \sbsalt{0.000} \\
    Aya             & \sbsalt{0.445} & \sbsalt{0.355} & \sbsalt{0.356} & \sbsalt{0.186} & \sbsalt{0.183} \\
    \bottomrule
  \end{tabular}}
  \caption{\label{performance_summary}
  Mean performance of models across programming languages.}
\end{table}

\noindent We observe that GPT-4o and DeepSeek-Coder achieve strong results in Classes 4 and 5, with scores consistently exceeding 0.85 in Python, Java, and C++. Python shows top performance, with scores reaching above 0.88 in Class 5. For lower classes (0-2), models like GPT-3.5, WizardCoder, and Aya underperform, often scoring below 0.70, particularly in JavaScript and Ruby, where scores frequently drop under 0.65. Even in higher classes, JavaScript and Ruby show challenges, with Class 4 scores for most models not exceeding 0.75. This highlights the models' limitations in handling non-Python languages, particularly for lower classes and specific scripting languages. While every model's best scores are generated with English-Python pair, DeepSeek-Coder is the only exception with Chinese-Python. 

A detailed analysis and discussion is provided in Appendix \ref{app:other_pls}.

\section{Insights and Analysis}

Upon curating the \dataset{} benchmark and completing the model evaluations, we now present some key analyses and gained insights based on the obtained results.

\subsection{LLMs' Performance Analysis} \label{insight1}
We observe significant performance discrepancies among the models, as illustrated by Figure \ref{fig:mHumanEval_plots_1}. While closed-source models perform better, their reliance on proprietary pretraining data complicates definitive conclusions. As suggested by the Chinchilla scaling hypothesis \cite{hoffmann2022training}, their superior performance may result from a larger parameter count and extensive training tokens, possibly including diverse and rare languages.

Aya, fine-tuned for multiple natural languages but not specifically for code generation, has the lowest \texttt{Pass@1} score in English. However, low variability across language classes indicates that multilingual pretraining and fine-tuning enhances code generation across different NLs.

WizardCoder's poor performance in non-English languages is due to its fine-tuning on StarCoder \cite{li2023starcoder}, which is primarily pretrained on code and documentation with minimal non-English content. In contrast, DeepSeek performs well for mid-resource languages but struggles with low-resource ones. These results suggest that effective multilingual code generation requires multilingual pretraining and/or finetuning datasets.

\begin{table*}[t]
  \centering
  \resizebox{0.85\textwidth}{!}{
  \begin{tabular}{lcccccccc}
    \toprule
     & \textbf{GPT4o} & \textbf{GPT3.5} & \textbf{Aya} & \textbf{WizardCoder} & \textbf{Claude3.5} & \textbf{DeepSeek-Coder} & \textbf{LLaMA 3} & \textbf{CodeStral} \\
    \midrule
    \texttt{mHumanEval-mini}   & \sbsalt{.72} & \sbsalt{.44} & \sbsalt{.47} & \sbsalt{.12} & \sbsalt{.61} & \sbsalt{.57} & \sbsalt{.35} & \sbsalt{.15}   \\
    \texttt{mHumanEval-T500}   & \sbsalt{.87} & \sbsalt{.76} & \sbsalt{.6} & \sbsalt{.63} & \sbsalt{.86} & \sbsalt{.73} & \sbsalt{.56} & \sbsalt{.36}   \\
    \texttt{mHumanEval-R500}   & \sbsalt{.78} & \sbsalt{.53} & \sbsalt{.47} & \sbsalt{.16} & \sbsalt{.59} & \sbsalt{.63} & \sbsalt{.28} & \sbsalt{.17}   \\
    \texttt{mHumanEval-B500}   & \sbsalt{.48} & \sbsalt{.21} & \sbsalt{.42} & \sbsalt{.00} & \sbsalt{.31} & \sbsalt{.22} & \sbsalt{.11} & \sbsalt{.10}   \\
    \bottomrule
  \end{tabular}}
  \caption{\label{ablation}
  Comparison of different LLMs' based on \% in \textbf{\texttt{Pass@1}} metric on multiple subsets of \dataset{}.}
\end{table*}

\subsection{Performance based on Language Classes} \label{insight2}
While there are significant discrepancies among the models' performances, a key trend observed is a somewhat consistent performance decline as we move from high-resource to low-resource languages. This decline is not as pronounced for Claude and GPT-4o. However, it is quite substantial for others and exceptionally steep for WizardCoder and DeepSeek-Coder.

\subsection{Error Analysis} \label{insight3}
In our analysis of errors, we observe several unique issues. Notably, the models rarely fail to generate any code. Specifically, GPT4o and GPT3.5 generate code with almost no compilation issues. However, a significant number of errors arise from misunderstandings of the problem, resulting in code that addresses incorrect tasks. This issue primarily occurs because translated keywords (e.g., string, list) do not always retain identical meanings in the target language, as illustrated in Appendix \ref{misunderstand}.

Furthermore, the Aya model often uses identifiers or keywords from different languages, leading to compilation errors (Appendix \ref{m_keywords}). A recurring problem with DeepSeek-Coder and WizardCoder is the generation of nonsensical code, sometimes not even in \texttt{Python}, especially when prompted in a non-English language (Appendix \ref{garbage}).

\subsection{Ablation Study} We present results from a limited ablation study conducted on various subsets of \dataset{} as detailed in Table \ref{tab:mHumanEval}. This study incorporates two additional models including MetaAI's \href{https://llama.meta.com/llama3/}{LLaMA 3} (70B), and MistralAI's code-finetuned \href{https://mistral.ai/news/codestral/}{CodeStral} (22B) model.

As indicated by the results in Table \ref{ablation}, \texttt{mHumanEval-mini} serves as an effective preliminary test for evaluating a model's proficiency in code generation following multilingual prompts. Models fine-tuned on code but lacking multilingual exposure perform poorly, whereas base models with some multilingual exposure perform better. The three subsets of \dataset{} are curated by prompt quality: \texttt{mHumanEval-T500} includes prompts from language class 5, \texttt{mHumanEval-B500} from classes 0 or 1, and \texttt{mHumanEval-R500} is randomly selected. These results align with our findings in Sections \ref{insight1} and \ref{insight2}.

\section{Conclusion}

This study introduces \dataset{}, a comprehensive multilingual code generation benchmark for assessing LLMs across 204 languages. We curated high-quality prompts for each language and evaluated various models. Our analyses, including ablation studies, provided insights into LLMs' multilingual code-generation capabilities, addressing the RQs posed in Section \ref{intro}:

\paragraph{RQ1:} \texttt{How do the code generation capabilities of LLMs vary when prompts are provided in English, or other high-, mid-, and low-resource NLs?} 

\begin{resultbox}
    LLMs generally demonstrate optimal performance when prompted in English. For prompts in other languages, performance varies based on the language's resource level. High-resource languages tend to yield superior results compared to mid- and low-resource languages. The extent of performance variation is contingent upon the specific language of the prompt and the model's prior exposure and training in that language. This variation is likely influenced by the model's training data and the relative abundance of resources available for each language.
\end{resultbox}

\paragraph{RQ2:} \texttt{How does the performance of multilingual LLMs compare to specialized, fine-tuned Code LLMs in code generation tasks on the \dataset{} dataset?}

\begin{resultbox}
    While code-finetuned language models excel at generating code from English prompts, multilingual models demonstrate strong proficiency across various NLs. Notably, even without specific code fine-tuning for different NLs, they achieve decent results in code generation. This phenomenon suggests that multilingual models can generalize coding capabilities across NLs, leveraging their understanding of multiple NLs to support diverse linguistic contexts in programming.
\end{resultbox}

\noindent While we draw some insightful conclusions from curating and evaluating \dataset{}, to facilitate further research, we are making it publicly available. We plan to expand coverage to more NLs and PLs in future updates. Despite the high cost of human translation, we included human annotations for 15 NLs, including some low-resource and rare ones. Currently, our dataset includes 164 prompts per language, following the \he benchmark, with plans to increase this number. We will also explore strategies to enhance low-resource language performance, such as transfer learning and diverse training datasets. Comparative studies between general-purpose multilingual LLMs and specialized code LLMs will help optimize multilingual code generation.

\section*{Limitations}

We conducted primary evaluations on six LLMs, focusing on key performance metrics. Given the benchmark's extensive 33,456 prompts, the evaluation process is exceedingly costly. This cost is the primary reason why we adopted \textbf{\texttt{Pass@1}} as our evaluation metric, rather than more resource-intensive metrics like \textbf{\texttt{Pass@10}} or \textbf{\texttt{Pass@100}}. However, to ensure a thorough analysis, we incorporated additional models in our ablation study. In our next iteration, we plan to comprehensively evaluate all models across the entire benchmark. This future work aims to enhance the benchmark’s robustness and provide deeper insights into the performance of various LLMs in multilingual code generation.

\section*{Ethical Considerations}
The benchmark introduced in this paper, which focuses on analyzing code generation using large language models (LLMs), strictly adheres to the \href{https://www.aclweb.org/portal/content/acl-code-ethics}{ACL Ethics Policy}. Each prompt in \dataset{} was tested multiple times by different models, and none produced any malicious code. Although there can occasionally be garbage code snippets or similar issues, none have posed any threats to the system.

To ensure safety and reliability, we recommend executing code generated using prompts from \dataset{} in a contained virtual environment. This precaution helps prevent potential issues related to infinite execution loops and memory management. Running code in a safe environment can also stop problems like crashing the system or using too much memory. We believe and hope that researchers and practitioners can maintain a secure and controlled testing environment while utilizing \dataset{}. This approach ensures that users can confidently explore and innovate without risking system integrity.

\section*{Acknowledgments}
We would like to thank the human annotators and experts for their valuable time and effort; also George Mason’s \href{https://orc.gmu.edu/}{Office of Research Computing (ORC)} for providing the computing resources. 

Antonios Anastasopoulos is additionally supported by the National Science Foundation under award IIS-2327143 and benefited from resources provided through the Microsoft Accelerate Foundation Models Research (AFMR) grant program.

\bibliography{custom}

\newpage

\appendix

\onecolumn

\section{list of NLs and PLs in \dataset{}}
\label{app:list}
\dataset{} supports 204 NLs and 25 PLs. The \texttt{Expert} subset contains human annotation for 15 NLs.

\subsection{List of PLs}

Comparing PL support provided by most widely used existing benchmarks -

\begin{table*}[!h]
\centering
\small
\begin{tabular}{l|ccccc|c}
\toprule
 & \multicolumn{6}{c}{Benchmarks} \\
 & \texttt{HumanEval} & \texttt{MBPP} & \texttt{Babel} \texttt{Code} & \texttt{MultiPL-E} & \texttt{HumanEval-XL} & \texttt{mHumanEval} \\
\midrule
Python & \greentick & \greentick & \greentick & \greentick & \greentick & \greentick \\
Bash & \redx & \redx & \redx & \greentick & \redx & \greentick \\
C++ & \redx & \redx & \greentick & \greentick & \redx & \greentick \\
C\# & \redx & \redx & \greentick & \greentick & \greentick & \greentick \\
D & \redx & \redx & \redx & \greentick & \redx & \greentick \\
Go & \redx & \redx & \greentick & \greentick & \greentick & \greentick \\
Haskell & \redx & \redx & \greentick & \redx & \redx & \greentick \\
Java & \redx & \redx & \greentick & \greentick & \greentick & \greentick \\
JavaScript & \redx & \redx & \greentick & \greentick & \greentick & \greentick \\
Julia & \redx & \redx & \redx & \greentick & \redx & \greentick \\
Kotlin & \redx & \redx & \greentick & \redx & \greentick & \greentick \\
Lua & \redx & \redx & \redx & \greentick & \redx & \greentick \\
Perl & \redx & \redx & \redx & \greentick & \greentick & \greentick \\
PHP & \redx & \redx & \greentick & \greentick & \greentick & \greentick \\
R & \redx & \redx & \redx & \greentick & \redx & \greentick \\
Racket & \redx & \redx & \redx & \greentick & \redx & \greentick \\
Ruby & \redx & \redx & \greentick & \greentick & \greentick & \greentick \\
Rust & \redx & \redx & \greentick & \greentick & \redx & \greentick \\
Scala & \redx & \redx & \greentick & \greentick & \greentick & \greentick \\
Swift & \redx & \redx & \greentick & \greentick & \greentick & \greentick \\
TypeScript & \redx & \redx & \greentick & \greentick & \greentick & \greentick \\
MATLAB & \redx & \redx & \redx & \redx & \redx & \greentick \\
Visual Basic & \redx & \redx & \redx & \redx & \redx & \greentick \\
Fortran & \redx & \redx & \redx & \redx & \redx & \greentick \\
COBOL & \redx & \redx & \redx & \redx & \redx & \greentick \\
\bottomrule
\end{tabular}
\caption{Comparing popular benchmarks in terms of NL and PL coverage.}
\label{tab:benchmarks}
\end{table*}

\subsection{List of NLs: \dataset{}-Expert}
Prompts in these languages are generated using translations done by native speakers, followed by evaluations done by expert programmers.

\begin{table}[h]
    \centering
    \scalebox{0.8}{
    \begin{tabular}{ll}
        \toprule
        \textbf{Language} & \textbf{Class} \\
        \midrule
        English & 5 \\
        Spanish & 5 \\
        French & 5 \\
        Japanese & 5 \\
        Arabic & 5 \\
        Chinese & 5 \\
        Portuguese & 4 \\
        Italian & 4 \\
        Korean & 4 \\
        Hindi & 4 \\
        Bangla & 3 \\
        Swahili & 2 \\
        Zulu & 2 \\
        Telugu & 1 \\
        Sinhala & 0 \\
        \bottomrule
    \end{tabular}}
    \caption{NLs along with their classes in \dataset{}-Expert.}
    \label{tab:language_classifications}
\end{table}

\subsection{List of NLs: \dataset{}}

\begin{table*}[h]
    \centering
    \scalebox{0.95}{
    \begin{tabular}{ll|ll|ll|ll}
        \toprule
        \textbf{Language} & \textbf{Class} & \textbf{Language} & \textbf{Class} & \textbf{Language} & \textbf{Class} & \textbf{Language} & \textbf{Class} \\
        \midrule
        arb\_Arab & 5 & zsm\_Latn & 3 & gla\_Latn & 1 & tat\_Cyrl & 1 \\
        deu\_Latn & 5 & amh\_Ethi & 2 & guj\_Gujr & 1 & tel\_Telu & 1 \\
        eng\_Latn & 5 & gle\_Latn & 2 & hye\_Armn & 1 & tgk\_Cyrl & 1 \\
        fra\_Latn & 5 & hau\_Latn & 2 & ibo\_Latn & 1 & tpi\_Latn & 1 \\
        jpn\_Jpan & 5 & isl\_Latn & 2 & ilo\_Latn & 1 & tso\_Latn & 1 \\
        spa\_Latn & 5 & lao\_Laoo & 2 & jav\_Latn & 1 & tuk\_Latn & 1 \\
        zho\_Hans & 5 & mar\_Deva & 2 & kab\_Latn & 1 & tum\_Latn & 1 \\
        cat\_Latn & 4 & mlt\_Latn & 2 & kan\_Knda & 1 & twi\_Latn & 1 \\
        ces\_Latn & 4 & pan\_Guru & 2 & kas\_Arab & 1 & uig\_Arab & 1 \\
        eus\_Latn & 4 & san\_Deva & 2 & kas\_Deva & 1 & vec\_Latn & 1 \\
        fin\_Latn & 4 & swh\_Latn & 2 & khk\_Cyrl & 1 & war\_Latn & 1 \\
        hin\_Deva & 4 & tir\_Ethi & 2 & khm\_Khmr & 1 & ydd\_Hebr & 1 \\
        hrv\_Latn & 4 & tsn\_Latn & 2 & kik\_Latn & 1 & zho\_Hant & 1 \\
        hun\_Latn & 4 & wol\_Latn & 2 & kin\_Latn & 1 & awa\_Deva & 0 \\
        ita\_Latn & 4 & xho\_Latn & 2 & kir\_Cyrl & 1 & bam\_Latn & 0 \\
        kor\_Hang & 4 & yor\_Latn & 2 & kmr\_Latn & 1 & ban\_Latn & 0 \\
        nld\_Latn & 4 & zul\_Latn & 2 & lij\_Latn & 1 & bem\_Latn & 0 \\
        pes\_Arab & 4 & ace\_Arab & 1 & lim\_Latn & 1 & cjk\_Latn & 0 \\
        pol\_Latn & 4 & ace\_Latn & 1 & lin\_Latn & 1 & dyu\_Latn & 0 \\
        por\_Latn & 4 & acm\_Arab & 1 & lmo\_Latn & 1 & fon\_Latn & 0 \\
        rus\_Cyrl & 4 & acq\_Arab & 1 & ltg\_Latn & 1 & fuv\_Latn & 0 \\
        srp\_Cyrl & 4 & aeb\_Arab & 1 & ltz\_Latn & 1 & grn\_Latn & 0 \\
        swe\_Latn & 4 & ajp\_Arab & 1 & lug\_Latn & 1 & hat\_Latn & 0 \\
        tur\_Latn & 4 & aka\_Latn & 1 & mai\_Deva & 1 & hne\_Deva & 0 \\
        vie\_Latn & 4 & als\_Latn & 1 & mal\_Mlym & 1 & kac\_Latn & 0 \\
        afr\_Latn & 3 & apc\_Arab & 1 & min\_Arab & 1 & kam\_Latn & 0 \\
        arb\_Latn & 3 & ars\_Arab & 1 & min\_Latn & 1 & kbp\_Latn & 0 \\
        arz\_Arab & 3 & ary\_Arab & 1 & mkd\_Cyrl & 1 & kea\_Latn & 0 \\
        ben\_Beng & 3 & asm\_Beng & 1 & mri\_Latn & 1 & kmb\_Latn & 0 \\
        bos\_Latn & 3 & ast\_Latn & 1 & mya\_Mymr & 1 & knc\_Arab & 0 \\
        bul\_Cyrl & 3 & ayr\_Latn & 1 & nno\_Latn & 1 & knc\_Latn & 0 \\
        ceb\_Latn & 3 & azb\_Arab & 1 & nob\_Latn & 1 & kon\_Latn & 0 \\
        dan\_Latn & 3 & azj\_Latn & 1 & npi\_Deva & 1 & lua\_Latn & 0 \\
        ell\_Grek & 3 & bak\_Cyrl & 1 & oci\_Latn & 1 & luo\_Latn & 0 \\
        est\_Latn & 3 & bel\_Cyrl & 1 & ory\_Orya & 1 & lus\_Latn & 0 \\
        glg\_Latn & 3 & bho\_Deva & 1 & pag\_Latn & 1 & mag\_Deva & 0 \\
        heb\_Hebr & 3 & bjn\_Arab & 1 & pap\_Latn & 1 & mni\_Beng & 0 \\
        ind\_Latn & 3 & bjn\_Latn & 1 & pbt\_Arab & 1 & mos\_Latn & 0 \\
        kat\_Geor & 3 & bod\_Tibt & 1 & plt\_Latn & 1 & nso\_Latn & 0 \\
        kaz\_Cyrl & 3 & bug\_Latn & 1 & quy\_Latn & 1 & nus\_Latn & 0 \\
        lit\_Latn & 3 & ckb\_Arab & 1 & sag\_Latn & 1 & nya\_Latn & 0 \\
        lvs\_Latn & 3 & crh\_Latn & 1 & sat\_Olck & 1 & prs\_Arab & 0 \\
        ron\_Latn & 3 & cym\_Latn & 1 & scn\_Latn & 1 & run\_Latn & 0 \\
        slk\_Latn & 3 & dik\_Latn & 1 & smo\_Latn & 1 & shn\_Mymr & 0 \\
        slv\_Latn & 3 & dzo\_Tibt & 1 & sna\_Latn & 1 & sin\_Sinh & 0 \\
        tam\_Taml & 3 & epo\_Latn & 1 & snd\_Arab & 1 & sot\_Latn & 0 \\
        tgl\_Latn & 3 & ewe\_Latn & 1 & som\_Latn & 1 & taq\_Latn & 0 \\
        tha\_Thai & 3 & fao\_Latn & 1 & srd\_Latn & 1 & taq\_Tfng & 0 \\
        ukr\_Cyrl & 3 & fij\_Latn & 1 & ssw\_Latn & 1 & tzm\_Tfng & 0 \\
        urd\_Arab & 3 & fur\_Latn & 1 & sun\_Latn & 1 & umb\_Latn & 0 \\
        uzn\_Latn & 3 & gaz\_Latn & 1 & szl\_Latn & 1 & yue\_Hant & 0 \\
        \bottomrule
    \end{tabular}}
    \caption{All NLs and their classes included in \dataset{}.}
    \label{tab:all_nls_classes}
\end{table*}

\clearpage

\twocolumn

\section{Prompt Translation and Evaluation Algorithm}
\label{alg:eval}

The pseudocode version of the workflow, presented in Figure \ref{fig:diagram}.

\begin{algorithm}
\caption{Prompt Translation and Evaluation}
\label{alg:translation_evaluation}
\begin{algorithmic}[1]
\FOR{each extracted prompt from HumanEval}
    \FOR{each translation system}
        \FOR{each target language}
            \IF{the language is supported}
                \STATE generate \underline{5} translated candidate prompts
                \STATE do back translation 
                \STATE calculate \textit{BERT\_Score} and \textit{Comet\_Kiwi} for each
                \STATE take the average of the two
                \STATE pick the best prompt
            \ELSE
                \STATE do back translation
                \STATE calculate only \textit{BERT\_Score}
                \STATE pick the best prompt
            \ENDIF
        \ENDFOR
    \ENDFOR
\ENDFOR
\end{algorithmic}
\end{algorithm}

It describes how the originally extracted prompts go through 13 candidate translations and evaluation via \texttt{BERTScore} and \texttt{CometKiwi} to build the new sets of benchmarks in the target natural languages.

\section{Evaluation Metric 1: \texttt{BERTScore}}
\label{app:bs}

\texttt{BERTScore} uses pre-trained BERT embeddings to assess similarity between candidate and reference translations. For a candidate sentence \( C \) and a reference sentence \( R \), let \( E_C \) and \( E_R \) be the sets of BERT embeddings for tokens in \( C \) and \( R \), respectively. The similarity \( S(i, j) \) between tokens \( i \) and \( j \) is the cosine similarity of their embeddings:

\[
S(i, j) = \frac{e_{C_i} \cdot e_{R_j}}{\|e_{C_i}\| \|e_{R_j}\|}
\]

Precision \( P \), recall \( R \), and F1-score \( F1 \) are then:

\[
P = \frac{1}{|E_C|} \sum_{e_{C_i} \in E_C} \max_{e_{R_j} \in E_R} S(i, j)
\]

\[
R = \frac{1}{|E_R|} \sum_{e_{R_j} \in E_R} \max_{e_{C_i} \in E_C} S(j, i)
\]

\[
F1 = 2 \cdot \frac{P \cdot R}{P + R}
\]

Here, \( P \) and \( R \) denote precision and recall as average maximum similarities from candidate to reference and vice versa. The \( F1 \) score is their harmonic mean.

\section{Evaluation Metric 2: \texttt{CometKiwi}}
\label{app:ck}

\texttt{CometKiwi} (Knowledge Integration via Weighted Importance) evaluates translations without references, using human-judgment scores. Given source \( \mathbf{x} \) and candidate \( \mathbf{y} \), it maps these inputs to a quality score \( Q(\mathbf{x}, \mathbf{y}) \) using a neural network \( \mathcal{N} \) trained on human scores \( Q_{\text{human}}(\mathbf{x}, \mathbf{y}) \):

\[
Q(\mathbf{x}, \mathbf{y}) = f(\mathbf{E}_{\text{src}}(\mathbf{x}), \mathbf{E}_{\text{cand}}(\mathbf{y}), \mathbf{L}(\mathbf{x}, \mathbf{y}))
\]

where \( \mathbf{E}_{\text{src}} \) and \( \mathbf{E}_{\text{cand}} \) are embeddings for \( \mathbf{x} \) and \( \mathbf{y} \), and \( \mathbf{L} \) represents linguistic features. The function \( f \) is:

\[
f = \mathcal{N}(\mathbf{E}_{\text{src}}, \mathbf{E}_{\text{cand}}, \mathbf{L})
\]

The network \( \mathcal{N} \) minimizes the loss:

\[
\mathcal{L} = \frac{1}{N} \sum_{i=1}^{N} \left( Q(\mathbf{x}_i, \mathbf{y}_i) - Q_{\text{human}}(\mathbf{x}_i, \mathbf{y}_i) \right)^2
\]

where \( N \) is the sample size.

\section{Annotator Details}

As mentioned in Section \ref{sec:expert}, \dataset{}-Expert utilizes native-speaking volunteer translators for 15 NLs. Each translator was assigned 164 prompts, with no monetary compensation involved. The experts, also native speakers, possess backgrounds in Computer Science and/or Information Technology, complemented by substantial coding experience. Both translators and experts were carefully selected through a rigorous process, ensuring a diverse demographic representation. This methodological approach enhances the dataset's linguistic diversity and technical robustness across various cultural contexts.

\clearpage

\onecolumn

\section{Comparison of the Prompt Qualities by the 3 models vs \texttt{mHumanEval}}
\label{app:comparison}

\begin{figure*}[h!]
    \centering
    \includegraphics[width=0.8\textwidth]{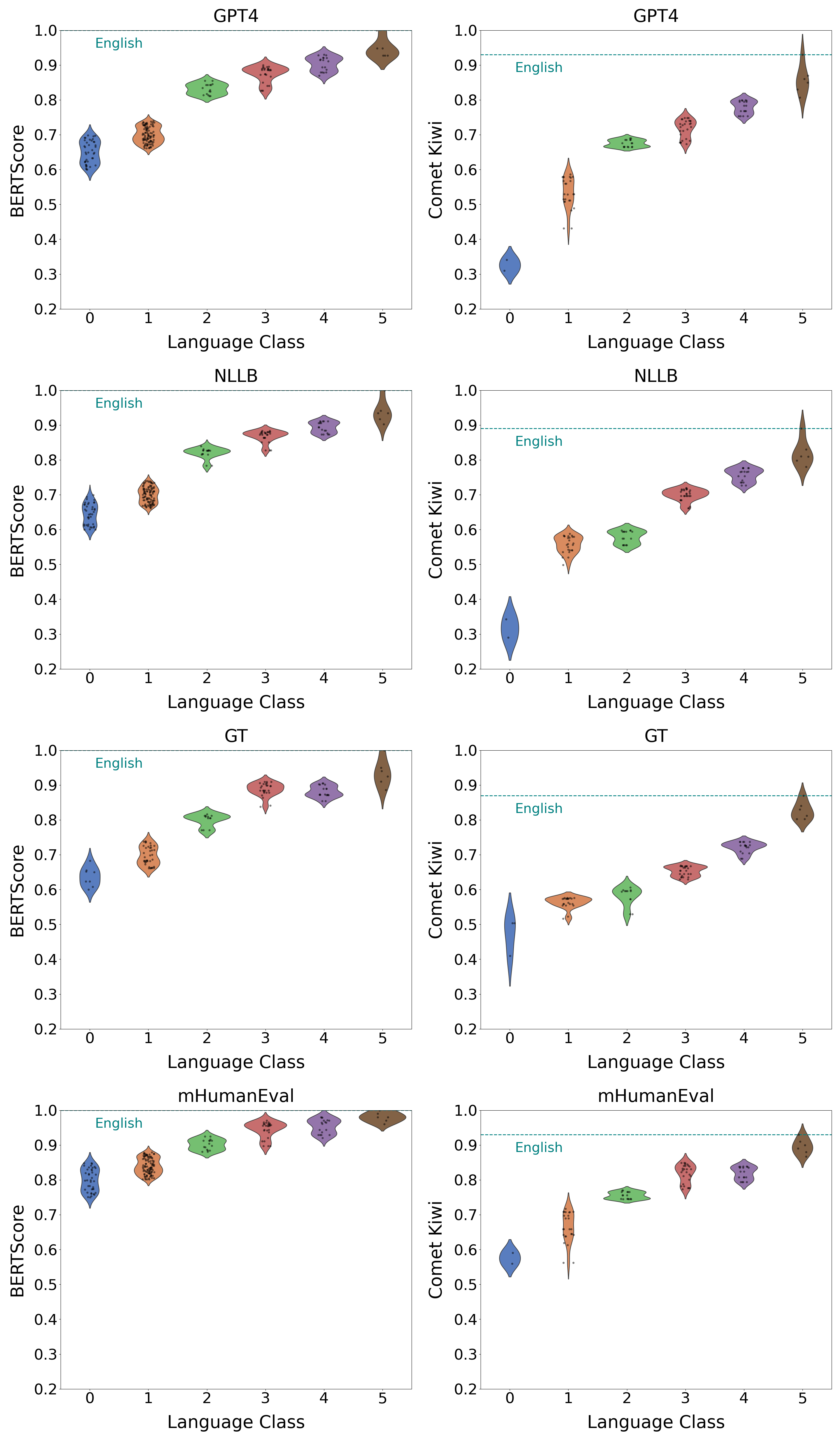}
    \caption{\label{fig:eval} Comparing the Machine Translation Quality for GPT4o, NLLB and Google Translator. The metrics used are \texttt{BERTScore} and \texttt{CometKiwi}. As shown in the figure, the prompts chosen for \dataset{} are better in quality upon choosing from 13 different candidates.}
\end{figure*}

\clearpage

\twocolumn
\section{Prompt Templates}
\label{app:prompts}

\subsection*{GPT4o and GPT3.5}

\begin{figure}[H]
    \centering
    \begin{minipage}{\textwidth}
        \begin{minted}{python}
    prompt = "Write a Python function for 
    the following: " + mHumanEval[i] + 
    " Ensure your response includes a 
    Python code block."
    
    messages=[
        {"role": "system", "content": 
        "You are a helpful assistant 
        trained to generate Python code.
        "},
        {"role": "user", "content": 
        prompt}
    ]
        
        \end{minted}
    \end{minipage}
    \caption{\label{fig:prompt}
    Prompt template - GPT4o and GPT3.5.}
\end{figure}


\subsection*{WizardCoder}


\begin{figure}[H]
    \centering
    \begin{minipage}{\textwidth}
        \begin{minted}{text}
    Below is an instruction that describes 
    a task. Write a response that 
    appropriately completes the request.
    
    ### Instruction:
    "mHumanEval[i]"
    
    ### Response:
        \end{minted}
    \end{minipage}
    \caption{\label{fig:prompt2}
    Prompt Template - WizardCoder}
\end{figure}


\subsection*{Aya}


\begin{figure}[H]
    \centering
    \begin{minipage}{\textwidth}
        \begin{minted}{python}
    messages = [{"role": "user", 
    "content": mHumanEval[i]}]
        \end{minted}
    \end{minipage}
    \caption{\label{fig:prompt3}
    Prompt Template - Aya.}
\end{figure}


\subsection*{Claude3.5}

\begin{figure}[H]
    \centering
    \begin{minipage}{0.9\textwidth}
        \begin{minted}{python}
    system="Write a Python Code snippet
    for the following: ",
    prompt = mHumanEval[i] + "Make sure 
    your response includes a code block."
    
    messages=[
        {"role": "user", 
        "content": prompt} 
    ]
        \end{minted}
    \end{minipage}
    \caption{\label{fig:code}
    Prompt template - Claude3-Opus.}
\end{figure}

\subsection*{LLaMA 3}

\begin{figure}[H]
    \centering
    \begin{minipage}{\textwidth}
        \begin{minted}{python}
    messages = [
        {"role": "system", "content": 
        "You are a helpful AI assistant, 
        who writes Python Code."},
        {"role": "user", "content": 
        mHumanEval[i]},
    ]
        \end{minted}
    \end{minipage}
    \caption{\label{fig:code}
    Prompt template - LLaMA 3.}
\end{figure}

\subsection*{DeepSeek-Coder}

\begin{figure}[H]
    \centering
    \begin{minipage}{\textwidth}
        \begin{minted}{python}
messages = 
    {"role": "user", "content": 
    "write a quick sort algorithm
    in python."}
        \end{minted}
    \end{minipage}
    \caption{\label{fig:code}
    Prompt template - DeepSeek-Coder.}
\end{figure}

\subsection*{CodeStral}

\begin{figure}[H]
    \centering
    \begin{minipage}{\textwidth}
        \begin{minted}{python}
    prompt = "Below is a programming 
    problem, paired with a language in 
    which the solution should be written.
    Write a solution in the provided that
    appropriately solves the programming 
    problem.

    ### Problem:
    
    mHumanEval[i]
    
    ### Language: python
    
    ### Solution: "
        \end{minted}
    \end{minipage}
    \caption{\label{fig:code}
    Prompt template - MagiCoder.}
\end{figure}

\clearpage

\section{Error Analysis - Examples}
\label{app:error}

\subsection{Task Misunderstanding} \label{misunderstand}
GPT4o generates a code for \texttt{finding significant digits}, when it is prompted to generate a code for \texttt{detecting prime numbers} in \texttt{Zulu}. This is due to the reason of the translation of \texttt{'prime number'} in Zulu becomes \texttt{'Inombolo Eyinhloko'} which means \texttt{'primary numbers'}. The model interprets it as \texttt{'significant number'} instead.

\begin{figure}[!h]
    \centering
    \small
    \begin{minipage}{\textwidth}
        \begin{minted}{python}
def find_significant_digits(number):
    
    # Convert the number to a string to handle it
    num_str = str(number)
    
    # Remove any leading/trailing zeros and 
    # the decimal point
    significant_digits = ''.join([char for char 
    in num_str if char.isdigit()])
    
    return significant_digits

# Example usage
number = 123.456
significant_digits = find_significant_digits(number)
print(f"Significant digits in {number} are: 
{significant_digits}")
        \end{minted}
    \end{minipage}
    \caption{\label{fig:mint} GP4o generated Python Code to \texttt{find significant digits}, when prompted to write code to \texttt{detect prime numbers} in \texttt{Zulu}.}
\end{figure}

Issues like this persist, even sometimes when prompted multiple times.


\subsection{Multilingual Keyword Issues} \label{m_keywords} Some code snippets include \texttt{keywords} and \texttt{identifiers} in the target language as well, resulting in compilation errors. As shown in Figure \ref{fig:key}, when prompted in \texttt{Rundi}, to \texttt{find the largest number in a list}, Aya generated a piece of code, where keywords like \texttt{for} and \texttt{return} turned into \texttt{kuri} and \texttt{subiza}.

\begin{figure}[!h]
    \centering
    \small
    \begin{minipage}{\textwidth}
        \begin{minted}{python}
def igitigiri_cane_kuruta(imibare):
    namba not imibare:  
        subiza None
    cane = imibare[0]
    kuri imibare hagati imibare:
        namba imibare > cane:
            cane = imibare
    subiza cane
        \end{minted}
    \end{minipage}
    \caption{\label{fig:key} When prompted in \texttt{Rundi}, Aya generated Python code to \texttt{find the largest number in a list}, containing multiple \texttt{keywords} in \texttt{Rundi}, resulting in Compilation Error.}
\end{figure}

These kinds of errors are also persistent but the outputs differ with similar prompts. Still generate the same issue again and again.


\subsection{Garbage Results} \label{garbage}

\begin{figure}[!h]
    \centering
    \small
    \begin{minipage}{\textwidth}
        \begin{minted}{c}
#include <stdio.h>

int main() {
    int x = 10; x = x / (x - x); 
    printf("%d\n", x); 
    x = x * "Hello World!";
}
        \end{minted}
    \end{minipage}
    \caption{\label{fig:mint} When prompted in \texttt{Sinhala}, to \texttt{reverse a list}, WizardCoder generated a garbage code in C.}
\end{figure}

\section{Experimental Setup}

\subsection{Machine Translation}
GPT4o is accessed via API key, eliminating the need for GPU hours. Hyperparameter tuning is not conducted; instead, recommended values are utilized. The \texttt{max\_tokens} parameter is set to 1000, and the \texttt{temperature} is maintained at 0.7. Additionally, Google Translate is accessed through API key, and the NLLB model is employed using a single NVIDIA A100 GPU with 40 GB of memory.

\subsection{Code Generation}
GPT4o, GPT3.5, and Claude3-Opus are accessed through API keys, thereby negating the necessity for GPU hours. We adhere to the recommended hyperparameters without conducting hyperparameter searches. The \texttt{max\_tokens} parameter is set to 1000, and the \texttt{temperature} is maintained at 0.7.

For WizardCoder and Aya, we utilize the full precision (FP32) models without employing any quantized versions. These models are run on four NVIDIA A100 GPUs, each with 40 GB of memory. Hyperparameter settings are maintained as per the authors' recommendations without additional tuning.

For MagiCoder, LLaMA 3, and Phi-3-mini, the full precision (FP32) models are employed on a single NVIDIA A100 GPU with 40 GB of memory. Hyperparameter configurations are again set to the recommended values as specified by the authors.

\clearpage

\onecolumn

\section{Evaluation Results: \dataset{}-{PL}}
\label{app:other_pls}

We evaluate the six LLMs from Table \ref{tab:models} for all 204 NLs in four different PLs. More specifically, we evaluate them on four subsets of \dataset{} - \dataset{}-C++, \dataset{}-JAVA, \dataset{}-JavaScript, and \dataset{}-Ruby. The results with \dataset{}-Python are presented in Figure \ref{fig:mHumanEval_plots_1} and discussed in Section \ref{sec:eval}. The performance trend is similar to Python, as discussed in Section \ref{sec:eval}. However, the results are slightly worse than those of Python.

\subsection{\dataset{}-C++}

\begin{figure*}[!h]
    \centering
    \includegraphics[width=0.8\textwidth]{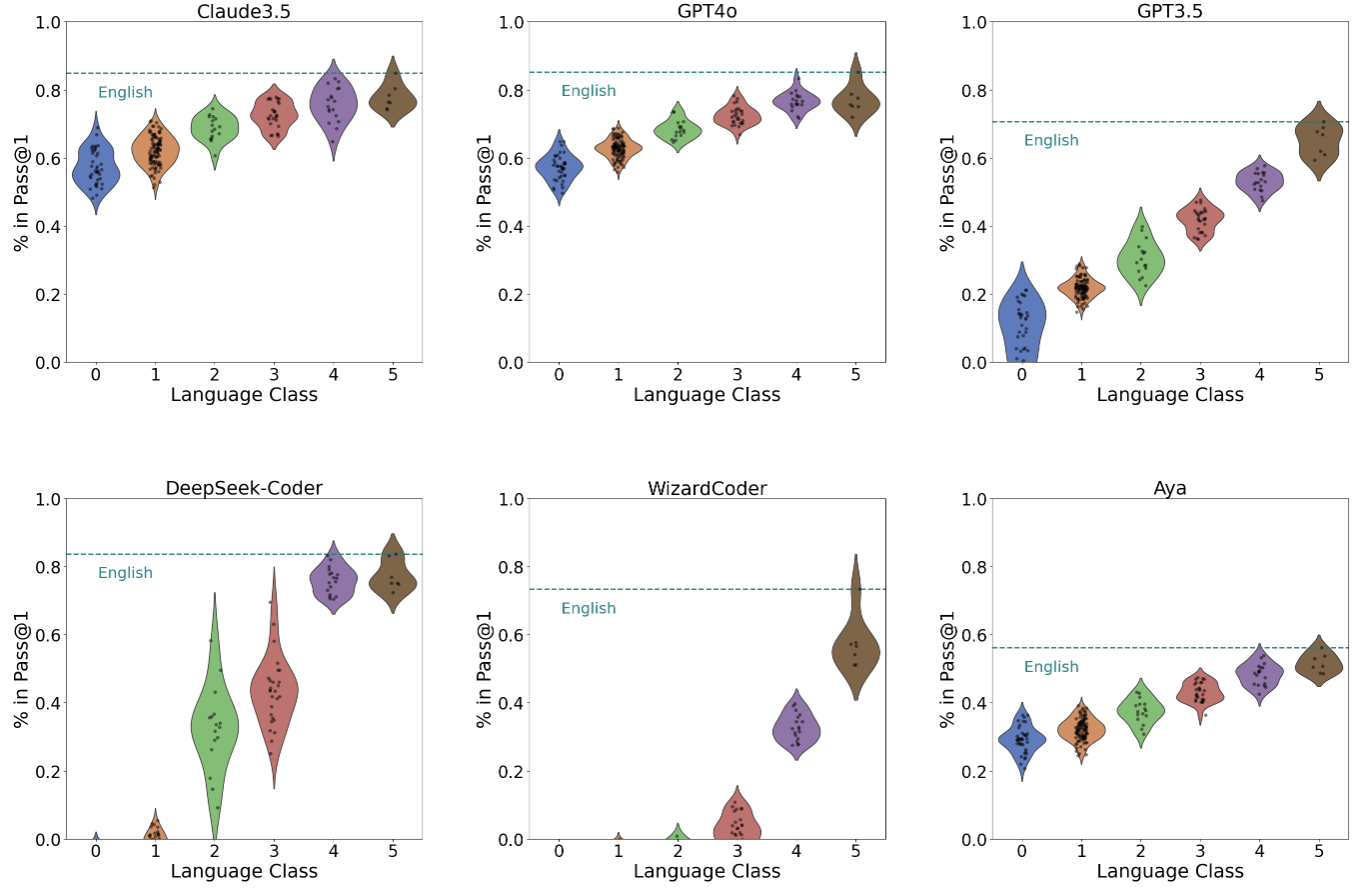}
    \caption{\label{fig:mHumanEval_plots2} Comparing model performances (\% in \textbf{\texttt{Pass@1}}) for the six models on mHumanEval-\textbf{C++}.}
\end{figure*}

\subsection{\dataset{}-JAVA}

\begin{figure*}[!h]
    \centering
    \includegraphics[width=0.8\textwidth]{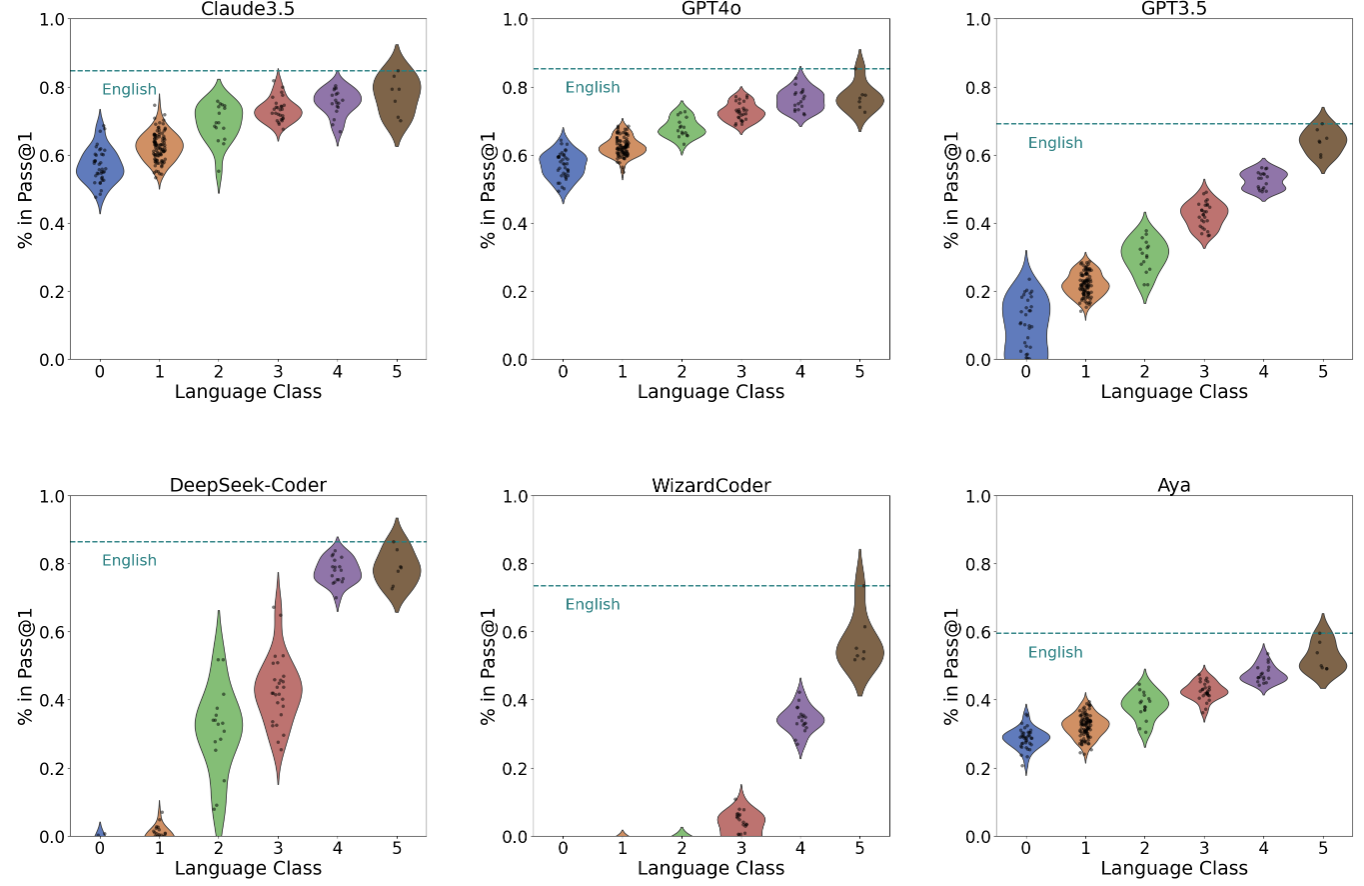}
    \caption{\label{fig:mHumanEval_plots3} Comparing model performances (\% in \textbf{\texttt{Pass@1}}) for the six models on mHumanEval-\textbf{JAVA}.}
\end{figure*}

\clearpage

\subsection{\dataset{}-JavaScript}

\begin{figure*}[!h]
    \centering
    \includegraphics[width=0.8\textwidth]{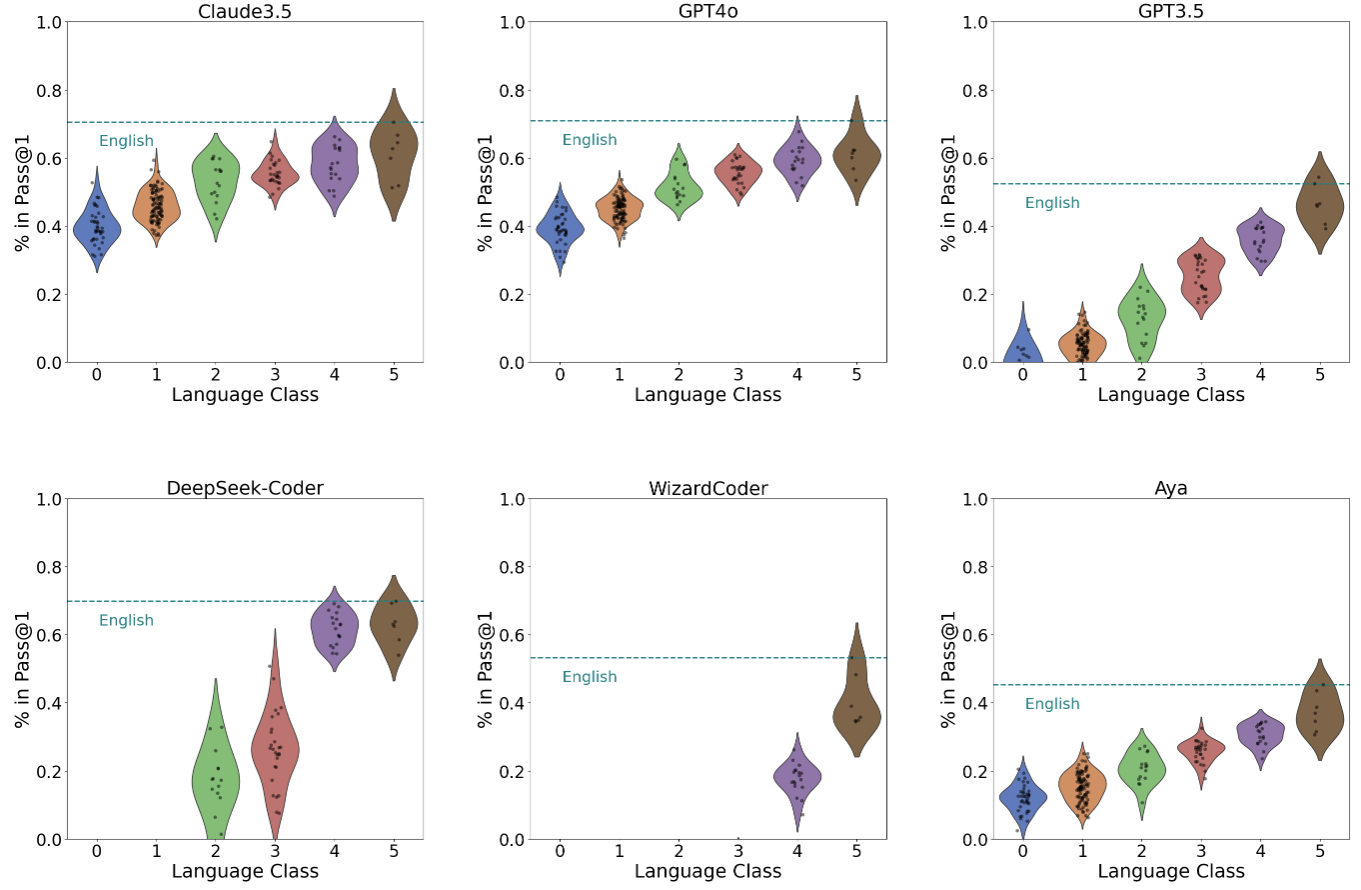}
    \caption{\label{fig:mHumanEval_plots4} Comparing model performances (\% in \textbf{\texttt{Pass@1}}) for the six models on mHumanEval-\textbf{JavaScript}.}
\end{figure*}

\subsection{\dataset{}-Ruby}

\begin{figure*}[!h]
    \centering
    \includegraphics[width=0.8\textwidth]{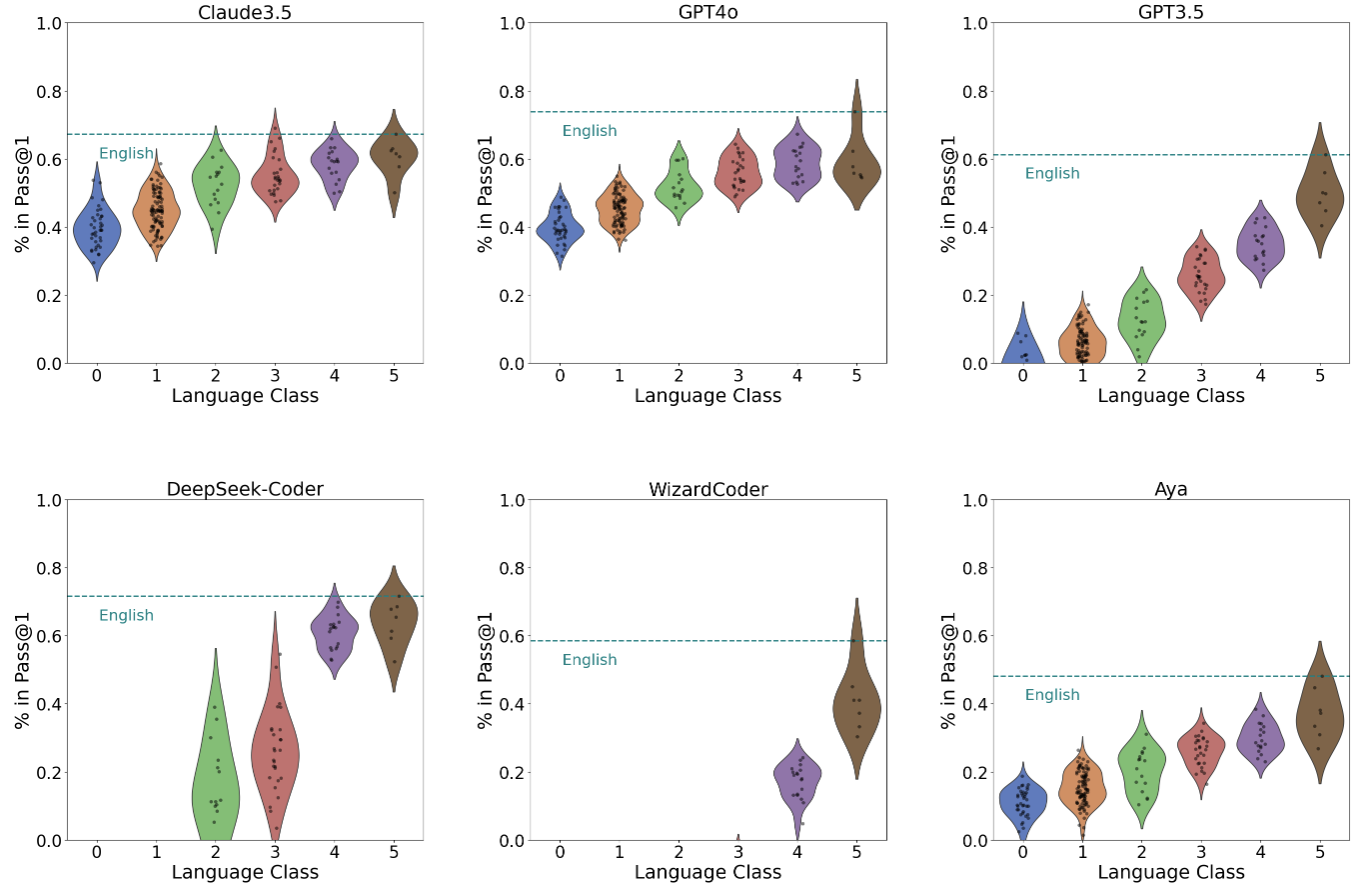}
    \caption{\label{fig:mHumanEval_plots5} Comparing model performances (\% in \textbf{\texttt{Pass@1}}) for the six models on mHumanEval-\textbf{Ruby}.}
\end{figure*}

\clearpage

\twocolumn

\subsection{Analyzing PL-specific results}

\paragraph{Performance Decline in Lower Classes (0-2)}
Models generally exhibit a noticeable performance decline in lower language classes, particularly Classes 0-2. Across all programming languages, scores in these classes fall well below the performance seen in Classes 4 and 5. This decline is especially pronounced in JavaScript and Ruby, where scores frequently drop to or near 0.000, suggesting these classes pose additional challenges.

\begin{table}[!h]
  \centering
  \resizebox{0.999\columnwidth}{!}{
  \begin{tabular}{lccccc}
    \toprule
    \textbf{Model} & \textbf{Python (C2)} & \textbf{Java (C2)} & \textbf{C++ (C2)} & \textbf{JavaScript (Cl1)} & \textbf{Ruby (C1)} \\
    \midrule
    GPT4o           & 0.600 & 0.590 & 0.591 & 0.000 & 0.000 \\
    GPT3.5          & 0.200 & 0.180 & 0.181 & 0.000 & 0.000 \\
    Claude3.5       & 0.620 & 0.600 & 0.601 & 0.478 & 0.473 \\
    DeepSeek-Coder  & 0.350 & 0.330 & 0.331 & 0.000 & 0.000 \\
    \bottomrule
  \end{tabular}}
  \caption{\label{lower_class_challenge}
  Performance of models in lower classes (0-2) across programming languages, with pronounced drops, particularly in JavaScript and Ruby.}
\end{table}

\paragraph{General Trends Across Language Classes}
In Classes 4 and 5, GPT-4 and Claude3.5 achieve high scores, often exceeding 0.85 in Python and Java. Python consistently demonstrates the highest scores, especially in Class 5, where models like GPT-4 and DeepSeek-Coder surpass 0.88. However, in Classes 0-3, performance drops across all models, particularly in JavaScript and Ruby, where scores frequently fall below 0.65.

\begin{table}[!h]
  \centering
  \resizebox{0.95\columnwidth}{!}{
  \begin{tabular}{lccccc}
    \toprule
    \textbf{Class} & \textbf{Python} & \textbf{Java} & \textbf{C++} & \textbf{JavaScript} & \textbf{Ruby} \\
    \midrule
    Class 5         & 0.880 & 0.850 & 0.852 & 0.650 & 0.653 \\
    Class 4         & 0.860 & 0.830 & 0.832 & 0.640 & 0.643 \\
    Class 3         & 0.750 & 0.720 & 0.721 & 0.530 & 0.533 \\
    Class 2         & 0.620 & 0.600 & 0.601 & 0.420 & 0.423 \\
    \bottomrule
  \end{tabular}}
  \caption{\label{class_performance}
  General model performance across language classes, highlighting high scores in Classes 4 and 5 and lower scores in Classes 0-3, particularly in JavaScript and Ruby.}
\end{table}

\paragraph{Underperformance of WizardCoder and Aya in JavaScript and Ruby Across Classes}
WizardCoder and Aya consistently struggle across all language classes in JavaScript and Ruby. In Classes 0-3, their scores frequently reach 0.000, underscoring limitations in handling these scripting languages regardless of language class.

\begin{table}[!h]
  \centering
  \resizebox{0.95\columnwidth}{!}{
  \begin{tabular}{lccccc}
    \toprule
    \textbf{Model} & \textbf{Class 5} & \textbf{Class 4} & \textbf{Class 3} & \textbf{Class 2} & \textbf{Class 1} \\
    \midrule
    WizardCoder (JavaScript) & 0.000 & 0.000 & 0.000 & 0.000 & 0.000 \\
    Aya (JavaScript)         & 0.186 & 0.165 & 0.143 & 0.120 & 0.100 \\
    WizardCoder (Ruby)       & 0.000 & 0.000 & 0.000 & 0.000 & 0.000 \\
    Aya (Ruby)               & 0.183 & 0.160 & 0.138 & 0.115 & 0.090 \\
    \bottomrule
  \end{tabular}}
  \caption{\label{wizard_aya_classes}
  Underperformance of WizardCoder and Aya in JavaScript and Ruby across language classes, with scores at 0.000 for WizardCoder across all classes.}
\end{table}

\paragraph{Mixed Adaptability of DeepSeek-Coder Across Language Classes}
DeepSeek-Coder shows moderate scores in Python for higher classes (Classes 4 and 5) but drops to 0.000 in lower classes, particularly in JavaScript and Ruby, highlighting issues with adaptability across classes.

\begin{table}[!h]
  \centering
  \resizebox{0.95\columnwidth}{!}{
  \begin{tabular}{lccccc}
    \toprule
    \textbf{Language Class} & \textbf{Python} & \textbf{Java} & \textbf{C++} & \textbf{JavaScript} & \textbf{Ruby} \\
    \midrule
    Class 5                 & 0.880 & 0.850 & 0.852 & 0.000 & 0.000 \\
    Class 4                 & 0.860 & 0.830 & 0.832 & 0.000 & 0.000 \\
    Class 3                 & 0.500 & 0.480 & 0.482 & 0.000 & 0.000 \\
    \bottomrule
  \end{tabular}}
  \caption{\label{deepseek_class}
  DeepSeek-Coder’s performance across language classes, illustrating high scores in Python and Java in Classes 4 and 5, but collapsing to 0.000 in JavaScript and Ruby.}
\end{table}

\paragraph{Claude3.5’s Stable Performance Across Language Classes}
Claude3.5 consistently scores above 0.477 across all languages and classes, indicating versatility and robust adaptability across different language classes and programming languages.

\begin{table}[!h]
  \centering
  \resizebox{0.95\columnwidth}{!}{
  \begin{tabular}{lccccc}
    \toprule
    \textbf{Language Class} & \textbf{Python} & \textbf{Java} & \textbf{C++} & \textbf{JavaScript} & \textbf{Ruby} \\
    \midrule
    Class 5                 & 0.880 & 0.850 & 0.852 & 0.483 & 0.477 \\
    Class 4                 & 0.860 & 0.830 & 0.832 & 0.480 & 0.475 \\
    Class 3                 & 0.750 & 0.720 & 0.721 & 0.480 & 0.475 \\
    Class 2                 & 0.620 & 0.600 & 0.601 & 0.478 & 0.473 \\
    \bottomrule
  \end{tabular}}
  \caption{\label{claude_class}
  Claude3.5’s consistent performance across language classes and programming languages, with scores remaining stable above 0.477.}
\end{table}

\paragraph{Implications for Future Model Development}
The significant underperformance in JavaScript and Ruby across language classes indicates a need for enhanced training in scripting languages. Models like GPT-4 and Claude3.5 excel in higher classes, particularly in Python and Java, but gaps in lower classes and scripting languages suggest a focus on diversifying training data to boost adaptability.

\clearpage

\onecolumn
\section{Evaluating Prompt Translation by GPT4}
\label{sec:mt_gpt4}

\begin{table*}[!h]
\centering
\resizebox{0.9\textwidth}{!}{
\begin{tabular}{lccc|lccc}
\toprule
\textbf{Language} & \textbf{Class} & \textbf{\texttt{BERTScore}} & \textbf{\texttt{CometKiwi}} & \textbf{Language} & \textbf{Class} & \textbf{\texttt{BERTScore}} & \textbf{\texttt{CometKiwi}} \\
\midrule
\cellcolor{class5}arb\_Arab & \cellcolor{class5}5 & \cellcolor{class5}{0.927} & \cellcolor{class5}{0.807} & \cellcolor{class3}tha\_Thai & \cellcolor{class3}3 & \cellcolor{class3}{0.874} & \cellcolor{class3}{0.749} \\
\cellcolor{class5}deu\_Latn & \cellcolor{class5}5 & \cellcolor{class5}{0.948} & \cellcolor{class5}{0.826} & \cellcolor{class3}ukr\_Cyrl & \cellcolor{class3}3 & \cellcolor{class3}{0.872} & \cellcolor{class3}{0.722} \\
\cellcolor{class5}eng\_Latn & \cellcolor{class5}5 & \cellcolor{class5}{1.000} & \cellcolor{class5}{0.930} & \cellcolor{class3}urd\_Arab & \cellcolor{class3}3 & \cellcolor{class3}{0.841} & \cellcolor{class3}{0.682} \\
\cellcolor{class5}fra\_Latn & \cellcolor{class5}5 & \cellcolor{class5}{0.927} & \cellcolor{class5}{0.807} & \cellcolor{class3}uzn\_Latn & \cellcolor{class3}3 & \cellcolor{class3}{0.885} & \cellcolor{class3}{0.740} \\
\cellcolor{class5}jpn\_Jpan & \cellcolor{class5}5 & \cellcolor{class5}{0.948} & \cellcolor{class5}{0.807} & \cellcolor{class3}zsm\_Latn & \cellcolor{class3}3 & \cellcolor{class3}{0.890} & \cellcolor{class3}{0.711} \\
\cellcolor{class5}spa\_Latn & \cellcolor{class5}5 & \cellcolor{class5}{0.927} & \cellcolor{class5}{0.839} & \cellcolor{class2}amh\_Ethi & \cellcolor{class2}2 & \cellcolor{class2}{0.825} & \cellcolor{class2}{0.690} \\
\cellcolor{class5}zho\_Hans & \cellcolor{class5}5 & \cellcolor{class5}{0.921} & \cellcolor{class5}{0.784} & \cellcolor{class2}gle\_Latn & \cellcolor{class2}2 & \cellcolor{class2}{0.824} & \cellcolor{class2}{0.666} \\
\bottomrule
\end{tabular}}
\caption{Evaluating the quality of machine translation by GPT4 using \texttt{BERTScore} and \texttt{CometKiwi}. The languages are given as Flores-200 codes.}
\label{tab:mt_gpt4}
\end{table*}

\begin{table*}[!h]
\centering
\small
\resizebox{0.67\textwidth}{!}{
\begin{tabular}{lcc|lcc}
\toprule
\textbf{Language} & \textbf{Class} & \textbf{\texttt{BERTScore}}  & \textbf{Language} & \textbf{Class} & \textbf{\texttt{BERTScore}} \\
\midrule
\cellcolor{class1}ace\_Arab & \cellcolor{class1}1 & \cellcolor{class1}{0.666} & \cellcolor{class1}quy\_Latn & \cellcolor{class1}1 & \cellcolor{class1}{0.722} \\
\cellcolor{class1}ace\_Latn & \cellcolor{class1}1 & \cellcolor{class1}{0.719} & \cellcolor{class1}sag\_Latn & \cellcolor{class1}1 & \cellcolor{class1}{0.735} \\
\cellcolor{class1}acm\_Arab & \cellcolor{class1}1 & \cellcolor{class1}{0.680} & \cellcolor{class1}sat\_Olck & \cellcolor{class1}1 & \cellcolor{class1}{0.717} \\
\cellcolor{class1}acq\_Arab & \cellcolor{class1}1 & \cellcolor{class1}{0.714} & \cellcolor{class1}scn\_Latn & \cellcolor{class1}1 & \cellcolor{class1}{0.730} \\
\cellcolor{class1}aeb\_Arab & \cellcolor{class1}1 & \cellcolor{class1}{0.664} & \cellcolor{class1}smo\_Latn & \cellcolor{class1}1 & \cellcolor{class1}{0.738} \\
\cellcolor{class1}ajp\_Arab & \cellcolor{class1}1 & \cellcolor{class1}{0.675} & \cellcolor{class1}sna\_Latn & \cellcolor{class1}1 & \cellcolor{class1}{0.679} \\
\cellcolor{class1}aka\_Latn & \cellcolor{class1}1 & \cellcolor{class1}{0.683} & \cellcolor{class1}srd\_Latn & \cellcolor{class1}1 & \cellcolor{class1}{0.705} \\
\bottomrule
\end{tabular}}
\caption{Evaluating the quality of machine translation by GPT4 using \texttt{BERTScore}. These languages are not supported by \texttt{CometKiwi}. The languages are given as Flores-200 codes.}
\label{tab:mt_gpt4_2}
\end{table*}

\newpage

\section{Evaluating Prompt Translation by NLLB}
\label{sec:mt_nllb}

\begin{table*}[!h]
\centering
\resizebox{0.9\textwidth}{!}{
\begin{tabular}{lccc|lccc}
\toprule
\textbf{Language} & \textbf{Class} & \textbf{\texttt{BERTScore}} & \textbf{\texttt{CometKiwi}} & \textbf{Language} & \textbf{Class} & \textbf{\texttt{BERTScore}} & \textbf{\texttt{CometKiwi}} \\
\midrule
\cellcolor{class5}arb\_Arab & \cellcolor{class5}5 & \cellcolor{class5}{0.941} & \cellcolor{class5}{0.798} & \cellcolor{class3}tha\_Thai & \cellcolor{class3}3 & \cellcolor{class3}{0.881} & \cellcolor{class3}{0.704} \\
\cellcolor{class5}deu\_Latn & \cellcolor{class5}5 & \cellcolor{class5}{0.902} & \cellcolor{class5}{0.809} & \cellcolor{class3}ukr\_Cyrl & \cellcolor{class3}3 & \cellcolor{class3}{0.883} & \cellcolor{class3}{0.685} \\
\cellcolor{class5}eng\_Latn & \cellcolor{class5}5 & \cellcolor{class5}{1.000} & \cellcolor{class5}{0.910} & \cellcolor{class3}urd\_Arab & \cellcolor{class3}3 & \cellcolor{class3}{0.872} & \cellcolor{class3}{0.697} \\
\cellcolor{class5}fra\_Latn & \cellcolor{class5}5 & \cellcolor{class5}{0.917} & \cellcolor{class5}{0.787} & \cellcolor{class3}uzn\_Latn & \cellcolor{class3}3 & \cellcolor{class3}{0.875} & \cellcolor{class3}{0.697} \\
\cellcolor{class5}jpn\_Jpan & \cellcolor{class5}5 & \cellcolor{class5}{0.935} & \cellcolor{class5}{0.807} & \cellcolor{class3}zsm\_Latn & \cellcolor{class3}3 & \cellcolor{class3}{0.864} & \cellcolor{class3}{0.697} \\
\cellcolor{class5}spa\_Latn & \cellcolor{class5}5 & \cellcolor{class5}{0.935} & \cellcolor{class5}{0.809} & \cellcolor{class2}amh\_Ethi & \cellcolor{class2}2 & \cellcolor{class2}{0.817} & \cellcolor{class2}{0.597} \\
\cellcolor{class5}zho\_Hans & \cellcolor{class5}5 & \cellcolor{class5}{0.911} & \cellcolor{class5}{0.831} & \cellcolor{class2}gle\_Latn & \cellcolor{class2}2 & \cellcolor{class2}{0.816} & \cellcolor{class2}{0.555} \\
\bottomrule
\end{tabular}}
\caption{Evaluating the quality of machine translation by NLLB using \texttt{BERTScore} and \texttt{CometKiwi}. The languages are given as Flores-200 codes.}
\label{tab:mt_nllb}
\end{table*}

\newpage

\begin{table*}[!h]
\centering
\small
\resizebox{0.67\textwidth}{!}{
\begin{tabular}{lcc|lcc}
\toprule
\textbf{Language} & \textbf{Class} & \textbf{\texttt{BERTScore}}  & \textbf{Language} & \textbf{Class} & \textbf{\texttt{BERTScore}} \\
\midrule
\cellcolor{class1}ace\_Arab & \cellcolor{class1}1 & \cellcolor{class1}{0.672} & \cellcolor{class1}quy\_Latn & \cellcolor{class1}1 & \cellcolor{class1}{0.700} \\
\cellcolor{class1}ace\_Latn & \cellcolor{class1}1 & \cellcolor{class1}{0.716} & \cellcolor{class1}sag\_Latn & \cellcolor{class1}1 & \cellcolor{class1}{0.702} \\
\cellcolor{class1}acm\_Arab & \cellcolor{class1}1 & \cellcolor{class1}{0.664} & \cellcolor{class1}sat\_Olck & \cellcolor{class1}1 & \cellcolor{class1}{0.681} \\
\cellcolor{class1}acq\_Arab & \cellcolor{class1}1 & \cellcolor{class1}{0.669} & \cellcolor{class1}scn\_Latn & \cellcolor{class1}1 & \cellcolor{class1}{0.668} \\
\cellcolor{class1}aeb\_Arab & \cellcolor{class1}1 & \cellcolor{class1}{0.664} & \cellcolor{class1}smo\_Latn & \cellcolor{class1}1 & \cellcolor{class1}{0.739} \\
\cellcolor{class1}ajp\_Arab & \cellcolor{class1}1 & \cellcolor{class1}{0.690} & \cellcolor{class1}sna\_Latn & \cellcolor{class1}1 & \cellcolor{class1}{0.679} \\
\cellcolor{class1}aka\_Latn & \cellcolor{class1}1 & \cellcolor{class1}{0.720} & \cellcolor{class1}srd\_Latn & \cellcolor{class1}1 & \cellcolor{class1}{0.694} \\
\bottomrule
\end{tabular}}
\caption{Evaluating the quality of machine translation by NLLB using \texttt{BERTScore}. These languages are not supported by \texttt{CometKiwi}. The languages are given as Flores-200 codes.}
\label{tab:mt_nllb}
\end{table*}

\newpage

\section{Evaluating Prompt Translation by Google Translate}
\label{sec:mt_gt}

\begin{table*}[!h]
\small
\centering
\begin{tabular}{lccc|lccc}
\toprule
\textbf{Language} & \textbf{Class} & \textbf{\texttt{BERTScore}} & \textbf{\texttt{CometKiwi}} & \textbf{Language} & \textbf{Class} & \textbf{\texttt{BERTScore}} & \textbf{\texttt{CometKiwi}} \\
\midrule
\cellcolor{class5}arb\_Arab & \cellcolor{class5}5 & \cellcolor{class5}{0.886} & \cellcolor{class5}{0.802} & \cellcolor{class2}lua\_Latn & \cellcolor{class2}2 & \cellcolor{class2}{0.825} & \cellcolor{class2}{0.645} \\
\cellcolor{class5}deu\_Latn & \cellcolor{class5}5 & \cellcolor{class5}{0.910} & \cellcolor{class5}{0.811} & \cellcolor{class2}ibo\_Latn & \cellcolor{class2}2 & \cellcolor{class2}{0.816} & \cellcolor{class2}{0.637} \\
\cellcolor{class5}eng\_Latn & \cellcolor{class5}5 & \cellcolor{class5}{1.000} & \cellcolor{class5}{0.950} & \cellcolor{class2}kin\_Latn & \cellcolor{class2}2 & \cellcolor{class2}{0.829} & \cellcolor{class2}{0.617} \\
\cellcolor{class5}fra\_Latn & \cellcolor{class5}5 & \cellcolor{class5}{0.924} & \cellcolor{class5}{0.802} & \cellcolor{class2}lao\_Laoo & \cellcolor{class2}2 & \cellcolor{class2}{0.829} & \cellcolor{class2}{0.645} \\
\cellcolor{class5}jpn\_Jpan & \cellcolor{class5}5 & \cellcolor{class5}{0.910} & \cellcolor{class5}{0.812} & \cellcolor{class2}lug\_Latn & \cellcolor{class2}2 & \cellcolor{class2}{0.835} & \cellcolor{class2}{0.632} \\
\cellcolor{class5}spa\_Latn & \cellcolor{class5}5 & \cellcolor{class5}{0.900} & \cellcolor{class5}{0.802} & \cellcolor{class2}luo\_Latn & \cellcolor{class2}2 & \cellcolor{class2}{0.832} & \cellcolor{class2}{0.617} \\
\cellcolor{class5}zho\_Hans & \cellcolor{class5}5 & \cellcolor{class5}{0.933} & \cellcolor{class5}{0.807} & \cellcolor{class2}mar\_Deva & \cellcolor{class2}2 & \cellcolor{class2}{0.836} & \cellcolor{class2}{0.637} \\
\bottomrule
\end{tabular}
\caption{Evaluating the quality of machine translation by Google Translator using \texttt{BERTScore} and \texttt{CometKiwi}. The languages are given as Flores-200 codes.}
\label{tab:mt_merged}
\end{table*}

\newpage

\section{Evaluating Prompt Quality in \dataset{}}
\label{mt_mhumaneval}

\begin{table*}[!h]
\centering
\resizebox{0.9\textwidth}{!}{
\begin{tabular}{lccc|lccc}
\toprule
\textbf{Language} & \textbf{Class} & \textbf{\texttt{BERTScore}} & \textbf{\texttt{CometKiwi}} & \textbf{Language} & \textbf{Class} & \textbf{\texttt{BERTScore}} & \textbf{\texttt{CometKiwi}} \\
\midrule
\cellcolor{class5}eng\_Latn & \cellcolor{class5}5 & \cellcolor{class5}1.000 & \cellcolor{class5}0.961 & \cellcolor{class3}urd\_Arab & \cellcolor{class3}3 & \cellcolor{class3}0.911 & \cellcolor{class3}0.782 \\
\cellcolor{class5}spa\_Latn & \cellcolor{class5}5 & \cellcolor{class5}0.98 & \cellcolor{class5}0.919 & \cellcolor{class3}bul\_Cyrl & \cellcolor{class3}3 & \cellcolor{class3}0.956 & \cellcolor{class3}0.777 \\
\cellcolor{class5}deu\_Latn & \cellcolor{class5}5 & \cellcolor{class5}0.99 & \cellcolor{class5}0.927 & \cellcolor{class3}ind\_Latn & \cellcolor{class3}3 & \cellcolor{class3}0.944 & \cellcolor{class3}0.777 \\
\cellcolor{class5}fra\_Latn & \cellcolor{class5}5 & \cellcolor{class5}0.98 & \cellcolor{class5}0.896 & \cellcolor{class3}tam\_Taml & \cellcolor{class3}3 & \cellcolor{class3}0.957 & \cellcolor{class3}0.777 \\
\cellcolor{class5}zho\_Hans & \cellcolor{class5}5 & \cellcolor{class5}0.96 & \cellcolor{class5}0.89 & \cellcolor{class3}heb\_Hebr & \cellcolor{class3}3 & \cellcolor{class3}0.965 & \cellcolor{class3}0.773 \\
\cellcolor{class5}jpn\_Jpan & \cellcolor{class5}5 & \cellcolor{class5}0.97 & \cellcolor{class5}0.88 & \cellcolor{class2}amh\_Ethi & \cellcolor{class2}2 & \cellcolor{class2}0.895 & \cellcolor{class2}0.77 \\
\cellcolor{class5}arb\_Arab & \cellcolor{class5}5 & \cellcolor{class5}0.96 & \cellcolor{class5}0.867 & \cellcolor{class2}mlt\_Latn & \cellcolor{class2}2 & \cellcolor{class2}0.904 & \cellcolor{class2}0.77 \\
\bottomrule
\end{tabular}}
\caption{Observing improved prompt quality in \dataset{} upon choosing the best ones from 13 candidates each, evaluated using \texttt{BERTScore} and \texttt{CometKiwi}. The languages are given as Flores-200 codes.}
\label{tab:mt_mhumaneval}
\end{table*}


\begin{table*}[!h]
\centering
\resizebox{0.58\textwidth}{!}{
\begin{tabular}{lcc|lcc}
\toprule
\textbf{Language} & \textbf{Class} & \textbf{\texttt{BERTScore}} & \textbf{Language} & \textbf{Class} & \textbf{\texttt{BERTScore}} \\
\midrule
\cellcolor{class1}ace\_Arab & \cellcolor{class1}1 & \cellcolor{class1}0.806 & \cellcolor{class1}quy\_Latn & \cellcolor{class1}1 & \cellcolor{class1}0.862 \\
\cellcolor{class1}ace\_Latn & \cellcolor{class1}1 & \cellcolor{class1}0.859 & \cellcolor{class1}sag\_Latn & \cellcolor{class1}1 & \cellcolor{class1}0.875 \\
\cellcolor{class1}acm\_Arab & \cellcolor{class1}1 & \cellcolor{class1}0.82 & \cellcolor{class1}sat\_Olck & \cellcolor{class1}1 & \cellcolor{class1}0.857 \\
\cellcolor{class1}acq\_Arab & \cellcolor{class1}1 & \cellcolor{class1}0.854 & \cellcolor{class1}scn\_Latn & \cellcolor{class1}1 & \cellcolor{class1}0.87 \\
\cellcolor{class1}aeb\_Arab & \cellcolor{class1}1 & \cellcolor{class1}0.804 & \cellcolor{class1}smo\_Latn & \cellcolor{class1}1 & \cellcolor{class1}0.878 \\
\cellcolor{class1}ajp\_Arab & \cellcolor{class1}1 & \cellcolor{class1}0.815 & \cellcolor{class1}sna\_Latn & \cellcolor{class1}1 & \cellcolor{class1}0.819 \\
\cellcolor{class1}aka\_Latn & \cellcolor{class1}1 & \cellcolor{class1}0.823 & \cellcolor{class1}srd\_Latn & \cellcolor{class1}1 & \cellcolor{class1}0.845 \\
\bottomrule
\end{tabular}}
\caption{Observing improved prompt quality in \dataset{} upon choosing the best ones from 13 candidates each, evaluated using \texttt{BERTScore}. These languages are not supported by \texttt{CometKiwi}. The languages are given as Flores-200 codes.}
\label{tab:mt_mhumaneval}
\end{table*}

\clearpage

\section{Evaluation Results on \textbf{\dataset{}}} \label{results}

\vspace{2cm}

\begin{table*}[!h]
    \centering
    \begin{tabular}{lccccccc}
        \toprule
        \textbf{Language} & \textbf{Class} & \textbf{Claude3.5} & \textbf{GPT4o} & \textbf{GPT3.5} & \textbf{DeepSeek-Coder} & \textbf{WizardCoder} & \textbf{Aya} \\
        \midrule
        \cellcolor{class5}arb\_Arab & \cellcolor{class5}5 & \sbsalt{0.831} & \sbsalt{0.846} & \sbsalt{0.719} & \sbsalt{0.859} & \sbsalt{0.650} & \sbsalt{0.590} \\
        \cellcolor{class5}deu\_Latn & \cellcolor{class5}5 & \sbsalt{0.846} & \sbsalt{0.833} & \sbsalt{0.730} & \sbsalt{0.863} & \sbsalt{0.670} & \sbsalt{0.620} \\
        \cellcolor{class5}eng\_Latn & \cellcolor{class5}5 & \sbsalt{0.938} & \sbsalt{0.910} & \sbsalt{0.770} & \sbsalt{0.902} & \sbsalt{0.800} & \sbsalt{0.650} \\
        \cellcolor{class5}fra\_Latn & \cellcolor{class5}5 & \sbsalt{0.835} & \sbsalt{0.850} & \sbsalt{0.693} & \sbsalt{0.849} & \sbsalt{0.650} & \sbsalt{0.608} \\
        \cellcolor{class5}jpn\_Jpan & \cellcolor{class5}5 & \sbsalt{0.896} & \sbsalt{0.868} & \sbsalt{0.757} & \sbsalt{0.849} & \sbsalt{0.670} & \sbsalt{0.609} \\
        \cellcolor{class5}spa\_Latn & \cellcolor{class5}5 & \sbsalt{0.880} & \sbsalt{0.852} & \sbsalt{0.759} & \sbsalt{0.854} & \sbsalt{0.610} & \sbsalt{0.609} \\
        \cellcolor{class5}zho\_Hans & \cellcolor{class5}5 & \sbsalt{0.838} & \sbsalt{0.810} & \sbsalt{0.720} & \sbsalt{0.933} & \sbsalt{0.590} & \sbsalt{0.570} \\
        \bottomrule
    \end{tabular}
    \caption{Comparing LLMs' performance (\% in \textbf{\texttt{Pass@1}}) on \texttt{mHumanEval} - Class 5 languages. The languages are given as Flores-200 codes.}
\end{table*}

\vspace{2cm}

\begin{table*}[!h]
    \centering
    \begin{tabular}{lccccccc}
        \toprule
        \textbf{Language} & \textbf{Class} & \textbf{Claude3.5} & \textbf{GPT4o} & \textbf{GPT3.5} & \textbf{DeepSeek-Coder} & \textbf{WizardCoder} & \textbf{Aya} \\
        \midrule
        \cellcolor{class4}cat\_Latn & \cellcolor{class4}4 & \sbsalt{0.764} & \sbsalt{0.832} & \sbsalt{0.613} & \sbsalt{0.827} & \sbsalt{0.420} & \sbsalt{0.584} \\
        \cellcolor{class4}ces\_Latn & \cellcolor{class4}4 & \sbsalt{0.908} & \sbsalt{0.837} & \sbsalt{0.649} & \sbsalt{0.883} & \sbsalt{0.390} & \sbsalt{0.591} \\
        \cellcolor{class4}eus\_Latn & \cellcolor{class4}4 & \sbsalt{0.880} & \sbsalt{0.884} & \sbsalt{0.617} & \sbsalt{0.902} & \sbsalt{0.480} & \sbsalt{0.599} \\
        \cellcolor{class4}fin\_Latn & \cellcolor{class4}4 & \sbsalt{0.857} & \sbsalt{0.882} & \sbsalt{0.611} & \sbsalt{0.882} & \sbsalt{0.390} & \sbsalt{0.565} \\
        \cellcolor{class4}hin\_Deva & \cellcolor{class4}4 & \sbsalt{0.854} & \sbsalt{0.859} & \sbsalt{0.600} & \sbsalt{0.872} & \sbsalt{0.480} & \sbsalt{0.572} \\
        \cellcolor{class4}hrv\_Latn & \cellcolor{class4}4 & \sbsalt{0.831} & \sbsalt{0.833} & \sbsalt{0.608} & \sbsalt{0.865} & \sbsalt{0.450} & \sbsalt{0.595} \\
        \cellcolor{class4}hun\_Latn & \cellcolor{class4}4 & \sbsalt{0.838} & \sbsalt{0.860} & \sbsalt{0.594} & \sbsalt{0.824} & \sbsalt{0.410} & \sbsalt{0.568} \\
\bottomrule
    \end{tabular}
    \caption{Comparing LLMs' performance (\% in \textbf{\texttt{Pass@1}}) on \texttt{mHumanEval} - Class 4 languages. The languages are given as Flores-200 codes.}
\end{table*}

\begin{table*}[!h]
    \centering
    \resizebox{0.9\textwidth}{!}{
    \begin{tabular}{lccccccc}
        \toprule
        \textbf{Language} & \textbf{Class} & \textbf{Claude3.5} & \textbf{GPT4o} & \textbf{GPT3.5} & \textbf{DeepSeek-Coder} & \textbf{WizardCoder} & \textbf{Aya} \\
        \midrule
        \cellcolor{class3}afr\_Latn & \cellcolor{class3}3 & \sbsalt{0.886} & \sbsalt{0.846} & \sbsalt{0.542} & \sbsalt{0.554} & \sbsalt{0.180} & \sbsalt{0.505} \\
        \cellcolor{class3}arb\_Latn & \cellcolor{class3}3 & \sbsalt{0.792} & \sbsalt{0.839} & \sbsalt{0.548} & \sbsalt{0.592} & \sbsalt{0.110} & \sbsalt{0.541} \\
        \cellcolor{class3}arz\_Arab & \cellcolor{class3}3 & \sbsalt{0.807} & \sbsalt{0.832} & \sbsalt{0.495} & \sbsalt{0.399} & \sbsalt{0.130} & \sbsalt{0.528} \\
        \cellcolor{class3}ben\_Beng & \cellcolor{class3}3 & \sbsalt{0.797} & \sbsalt{0.792} & \sbsalt{0.541} & \sbsalt{0.565} & \sbsalt{0.090} & \sbsalt{0.523} \\
        \cellcolor{class3}bos\_Latn & \cellcolor{class3}3 & \sbsalt{0.826} & \sbsalt{0.812} & \sbsalt{0.502} & \sbsalt{0.746} & \sbsalt{0.140} & \sbsalt{0.546} \\
        \cellcolor{class3}bul\_Cyrl & \cellcolor{class3}3 & \sbsalt{0.848} & \sbsalt{0.796} & \sbsalt{0.491} & \sbsalt{0.379} & \sbsalt{0.120} & \sbsalt{0.536} \\
        \cellcolor{class3}ceb\_Latn & \cellcolor{class3}3 & \sbsalt{0.850} & \sbsalt{0.827} & \sbsalt{0.499} & \sbsalt{0.473} & \sbsalt{0.150} & \sbsalt{0.479} \\
\bottomrule
    \end{tabular}}
    \caption{Comparing LLMs' performance (\% in \textbf{\texttt{Pass@1}}) on \texttt{mHumanEval} - Class 3 languages. The languages are given as Flores-200 codes.}
\end{table*}

\begin{table*}[!h]
    \centering
    \resizebox{0.9\textwidth}{!}{
    \begin{tabular}{lccccccc}
        \toprule
        \textbf{Language} & \textbf{Class} & \textbf{Claude3.5} & \textbf{GPT4o} & \textbf{GPT3.5} & \textbf{DeepSeek-Coder} & \textbf{WizardCoder} & \textbf{Aya} \\
        \midrule
        \cellcolor{class2}amh\_Ethi & \cellcolor{class2}2 & \sbsalt{0.765} & \sbsalt{0.742} & \sbsalt{0.373} & \sbsalt{0.214} & \sbsalt{0.020} & \sbsalt{0.454} \\
        \cellcolor{class2}gle\_Latn & \cellcolor{class2}2 & \sbsalt{0.753} & \sbsalt{0.748} & \sbsalt{0.466} & \sbsalt{0.425} & \sbsalt{0.010} & \sbsalt{0.449} \\
        \cellcolor{class2}hau\_Latn & \cellcolor{class2}2 & \sbsalt{0.670} & \sbsalt{0.739} & \sbsalt{0.431} & \sbsalt{0.382} & \sbsalt{0.070} & \sbsalt{0.447} \\
        \cellcolor{class2}isl\_Latn & \cellcolor{class2}2 & \sbsalt{0.795} & \sbsalt{0.770} & \sbsalt{0.419} & \sbsalt{0.606} & \sbsalt{0.030} & \sbsalt{0.439} \\
        \cellcolor{class2}lao\_Laoo & \cellcolor{class2}2 & \sbsalt{0.783} & \sbsalt{0.745} & \sbsalt{0.449} & \sbsalt{0.440} & \sbsalt{0.050} & \sbsalt{0.516} \\
        \cellcolor{class2}mar\_Deva & \cellcolor{class2}2 & \sbsalt{0.764} & \sbsalt{0.773} & \sbsalt{0.464} & \sbsalt{0.493} & \sbsalt{0.050} & \sbsalt{0.519} \\
        \cellcolor{class2}mlt\_Latn & \cellcolor{class2}2 & \sbsalt{0.826} & \sbsalt{0.790} & \sbsalt{0.348} & \sbsalt{0.184} & \sbsalt{0.020} & \sbsalt{0.439} \\
\bottomrule
    \end{tabular}}
    \caption{Comparing LLMs' performance (\% in \textbf{\texttt{Pass@1}}) on \texttt{mHumanEval} - Class 2 languages. The languages are given as Flores-200 codes.}
\end{table*}

\begin{table*}[!h]
    \centering
    \resizebox{0.47\textwidth}{!}{
    \begin{tabular}{lccccccc}
        \toprule
        \textbf{Language} & \textbf{Class} & \textbf{Claude3.5} & \textbf{GPT4o} & \textbf{GPT3.5} & \textbf{DeepSeek-Coder} & \textbf{WizardCoder} & \textbf{Aya} \\
        \midrule
        \cellcolor{class1}ace\_Arab & \cellcolor{class1}1 & \sbsalt{0.812} & \sbsalt{0.736} & \sbsalt{0.281} & \sbsalt{0.070} & \sbsalt{0.050} & \sbsalt{0.423} \\
        \cellcolor{class1}ace\_Latn & \cellcolor{class1}1 & \sbsalt{0.712} & \sbsalt{0.675} & \sbsalt{0.338} & \sbsalt{0.005} & \sbsalt{0.040} & \sbsalt{0.433} \\
        \cellcolor{class1}acm\_Arab & \cellcolor{class1}1 & \sbsalt{0.673} & \sbsalt{0.671} & \sbsalt{0.276} & \sbsalt{0.019} & \sbsalt{0.010} & \sbsalt{0.437} \\
        \cellcolor{class1}acq\_Arab & \cellcolor{class1}1 & \sbsalt{0.786} & \sbsalt{0.750} & \sbsalt{0.284} & \sbsalt{0.019} & \sbsalt{0.030} & \sbsalt{0.387} \\
        \cellcolor{class1}aeb\_Arab & \cellcolor{class1}1 & \sbsalt{0.716} & \sbsalt{0.739} & \sbsalt{0.324} & \sbsalt{0.036} & \sbsalt{0.010} & \sbsalt{0.413} \\
        \cellcolor{class1}ajp\_Arab & \cellcolor{class1}1 & \sbsalt{0.640} & \sbsalt{0.686} & \sbsalt{0.282} & \sbsalt{0.046} & \sbsalt{0.030} & \sbsalt{0.376} \\
        \cellcolor{class1}aka\_Latn & \cellcolor{class1}1 & \sbsalt{0.687} & \sbsalt{0.708} & \sbsalt{0.345} & \sbsalt{0.124} & \sbsalt{0.020} & \sbsalt{0.371} \\
\bottomrule
    \end{tabular}}
    \caption{Comparing LLMs' performance (\% in \textbf{\texttt{Pass@1}}) on \texttt{mHumanEval} - Class 1 languages. The languages are given as Flores-200 codes.}
\end{table*}

\clearpage

\begin{table*}[!h]
    \centering
    \resizebox{0.8\textwidth}{!}{
    \begin{tabular}{lccccccc}
        \toprule
        \textbf{Language} & \textbf{Class} & \textbf{Claude3.5} & \textbf{GPT4o} & \textbf{GPT3.5} & \textbf{DeepSeek-Coder} & \textbf{WizardCoder} & \textbf{Aya} \\
        \midrule
        \cellcolor{class0}awa\_Deva & \cellcolor{class0}0 & \sbsalt{0.653} & \sbsalt{0.628} & \sbsalt{0.191} & \sbsalt{0.033} & \sbsalt{0.020} & \sbsalt{0.353} \\
        \cellcolor{class0}bam\_Latn & \cellcolor{class0}0 & \sbsalt{0.645} & \sbsalt{0.634} & \sbsalt{0.268} & \sbsalt{0.081} & \sbsalt{0.010} & \sbsalt{0.410} \\
        \cellcolor{class0}ban\_Latn & \cellcolor{class0}0 & \sbsalt{0.639} & \sbsalt{0.641} & \sbsalt{0.285} & \sbsalt{0.060} & \sbsalt{0.010} & \sbsalt{0.398} \\
        \cellcolor{class0}bem\_Latn & \cellcolor{class0}0 & \sbsalt{0.675} & \sbsalt{0.654} & \sbsalt{0.308} & \sbsalt{0.000} & \sbsalt{0.000} & \sbsalt{0.415} \\
        \cellcolor{class0}cjk\_Latn & \cellcolor{class0}0 & \sbsalt{0.750} & \sbsalt{0.720} & \sbsalt{0.316} & \sbsalt{0.000} & \sbsalt{0.010} & \sbsalt{0.366} \\
        \cellcolor{class0}dyu\_Latn & \cellcolor{class0}0 & \sbsalt{0.620} & \sbsalt{0.636} & \sbsalt{0.039} & \sbsalt{0.000} & \sbsalt{0.010} & \sbsalt{0.367} \\
        \cellcolor{class0}fon\_Latn & \cellcolor{class0}0 & \sbsalt{0.719} & \sbsalt{0.658} & \sbsalt{0.072} & \sbsalt{0.016} & \sbsalt{0.000} & \sbsalt{0.396} \\
\bottomrule
    \end{tabular}}
    \caption{Comparing LLMs' performance (\% in \textbf{\texttt{Pass@1}}) on \texttt{mHumanEval} - Class 0 languages. The languages are given as Flores-200 codes.}
\end{table*}

\clearpage

\end{document}